\documentclass[acmlarge]{acmart}

\usepackage{booktabs}
\usepackage{multirow}
\usepackage{siunitx}

\DeclareMathOperator*{\argmin}{\arg\!\min}

\AtBeginDocument{%
  \providecommand\BibTeX{{%
    \normalfont B\kern-0.5em{\scshape i\kern-0.25em b}\kern-0.8em\TeX}}}

\setcopyright{acmcopyright}
\copyrightyear{2020}
\acmYear{2020}
\acmDOI{}

\begin{document}

\title{Adversarial Transferability in Wearable Sensor Systems}


\author{Ramesh Kumar Sah}
\affiliation{
  \institution{Washington State University}
  \streetaddress{P.O. Box 642752}
  \city{Pullman}
  \state{Washington}
  \postcode{99164-2752}
}
\email{ramesh.sah@wsu.edu}
\author{Hassan Ghasemzadeh}
\affiliation{
  \institution{Washington State University}
  \streetaddress{P.O. Box 642752}
  \city{Pullman}
  \state{Washington}
  \postcode{99164-2752}
}
\email{hassan.ghasemzadeh@wsu.edu}



\begin{abstract}
Machine learning is used for inference and decision making in wearable sensor systems. However, recent studies have found that machine learning algorithms are easily fooled by the addition of adversarial perturbations to their inputs. What is more interesting is that adversarial examples generated for one machine learning system is also effective against other systems. This property of adversarial examples is called {\it transferability}. In this work, we take the first stride in studying adversarial transferability in wearable sensor systems from the following perspectives: 1) transferability between machine learning systems, 2) transferability across subjects, 3) transferability across sensor body locations, and 4) transferability across datasets. We found strong untargeted transferability in most cases. Targeted attacks were less successful with success scores from $0\%$ to $80\%$. The transferability of adversarial examples depends on many factors such as the inclusion of data from all subjects, sensor body position, number of samples in the dataset, type of learning algorithm, and the distribution of source and target system dataset. The transferability of adversarial examples decreases sharply when the data distribution of the source and target system becomes more distinct. We also provide guidelines for the community for designing robust sensor systems.
\end{abstract}





\maketitle

\section{Introduction}
Human Activity Recognition (HAR) has a substantial footprint in the field of mobile and ubiquitous computing. Traditionally, systems designed for activity recognition utilized videos and images as inputs \cite{7518920}, but recently there has been a significant shift towards the use of inertial sensors such as accelerometer, gyroscope, and magnetometer \cite{lara_survey_2013}. Moreover, advances in sensor technology and detection algorithms allow for real-time and continuous detection of human activities with battery-powered devices of small form factor to be used in daily living situations \cite{kasteren}. This has propelled the use of wearable devices for human activity recognition in areas such as health monitoring, patients rehabilitation, athlete performance assessment, and medicine adherence \cite{10.1145/3314399, 10.1145/3351281, 10.1145/3130927}.

\begin{figure}[!tbh]
    \centering
    \includegraphics[width=\linewidth]{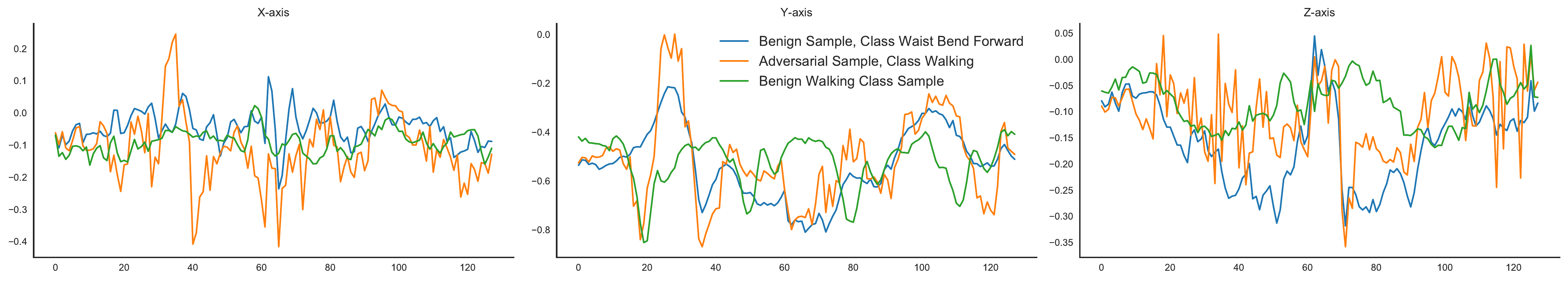}
    \caption{3-axial graphs for a benign sample take from the dataset, a targeted adversarial sample computed using the Basic Iterative Method (BIM), and a benign sample for the target class. }
    \label{fig:benign_and_adv_sample}
\end{figure}

In the current context, machine learning algorithms have become an integral part of most HAR systems \cite{10.1145/3090076, 10.1145/3161174, 10.1145/3214277}. In particular, the state-of-the-art systems and commercial devices used for human activity recognition use machine learning algorithms to detect various bio-markers and joint movements to provide real-time continuous clinical assessments based on the sensor data. However, the exceptional performance of these machine learning algorithms is not without any shortcomings. Recent studies have found that an adversary can easily fool machine learning algorithms with the addition of carefully computed perturbation to their inputs \cite{adar, GoodFellow14, Biggio13, Szegedy14}. These perturbed inputs are referred to as \emph{adversarial examples}. Even the addition of a small amount of carefully computed perturbations to the benign inputs, as shown in Fig \ref{fig:benign_and_adv_sample}, can degrade the performance of machine learning systems significantly \cite{adar, Szegedy14, Papernot16, kurakin2018adversarial, athalye2018synthesizing}. What distinguishes adversarial perturbations from random noise is that adversarial examples are misclassified far more often than samples that have been perturbed by random noise, even if the magnitude of random noise is much larger compared to the adversarial perturbation \cite{Szegedy14}. 

\begin{figure}[!tbh]
    \centering
    \includegraphics[width=\linewidth]{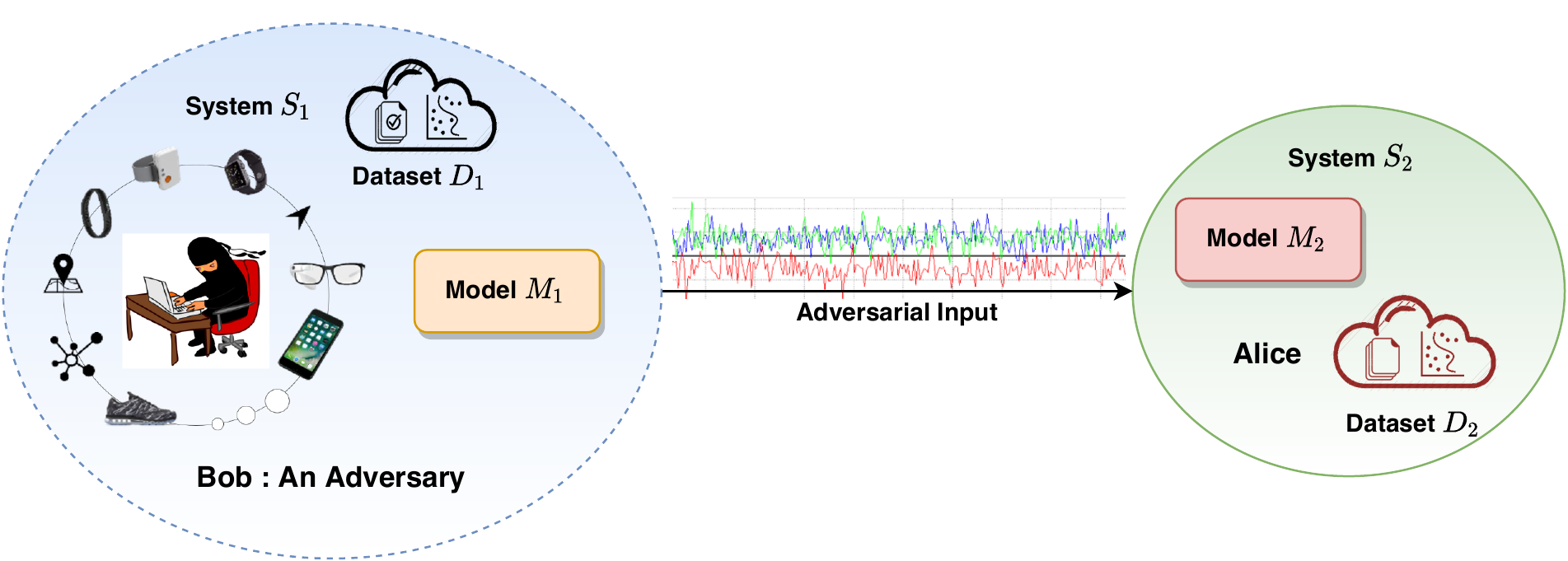}
    \caption{The operating scenario for transferability of adversarial examples. Here, Bob is an adversary with complete access to source system $S_1$ that constitutes souce model $M_1$ and source dataset $D_1$ and wants to attack Alice's system $S_2$ which contain target model $M_2$ trained on target dataset $D_2$. Bob computes adversarial examples using $M_1$ and transfers them to $M_2$ in the hope of fooling $M_2$ by exploiting the transferability of adversarial examples. Here, models $M_1$ and $M_2$ can use different learning algorithms and have different architectures and hyper-parameter values. Also, the source and target datasets can be distinct or same based on the type of adversarial transferability.}
    \label{fig:alice_bob}
\end{figure}

\subsection{Motivation}
Before we discuss our contributions, we first establish the threat model that is used in this work and answer why the transferability of adversarial examples is crucial to the discussion of robustness of wearable sensor systems. Assume that Bob, an adversary, has complete access to source system $S_1$ that contains source machine learning model $M_1$ trained using source dataset $D_1$ and Alice, a system administrator, wants to protect her system $S_2$ that uses target machine learning model $M_2$ trained on target dataset $D_2$ as shown in Fig. \ref{fig:alice_bob}. The only difference is that model $M_1$ is trained on dataset $D_1$ available freely, and Alice trains $M_2$ on the same dataset or some proprietary dataset $D_2$ only available to her. Also, the models $M_1$ and $M_2$ can use different learning algorithms and have different architectures and hyper-parameter values. Bob wants to attack Alice's system, such that $M_2$ is fooled in classifying inputs into wrong classes. Bob has access to Alice's model via an oracle, and hence can submit inputs and observe outputs. Bob can attack Alice's system $S_2$ in one of two ways. Bob can either compute adversarial examples using $M_1$ and $D_1$ and transfer them to $M_2$ in the hope of fooling $M_2$ or train a substitute model $M_S$ on dataset $D_S$ generated using the oracle and then use the substitute model to compute adversarial examples to fool $M_2$. In both of these cases, Bob tries to exploit the transferability property of adversarial examples to attack target system $S_2$. This is precisely the motivation of our work, in which we explore different types of adversarial transferability, inherent to wearable sensor systems, Bob can exploit to attack Alice's system. In the discussion, that follows we recognize Bob's system as \textit{Source System} and Alice's system as \textit{Target System}. For all four adversarial transferability modes we have discussed in this work, Bob has complete access to source system $S_1$, but can only query Alice's system target $S_2$ on inputs and observe outputs.

Adversarial transferability captures the ability of an adversarial attack against a machine learning system to be effective against other independently trained systems \cite{Demontis_why}. The transferability of adversarial examples was first examined in \cite{Szegedy14}, in which the authors studied adversarial transferability 1) between different machine learning models trained over the same dataset, and 2) between same or different machine learning models trained over disjoint subsets of a dataset. Motivated by the results of \cite{Demontis_why}, numerous studies have explored adversarial transferability for both test-time evasion attacks and training-time poisoning attacks \cite{adar, Demontis_why, Papernot16}. Furthermore, prior research has shown that adversarial examples can be generated in wearable sensor systems for human activity recognition \cite{adar}. However, the transferability of adversarial examples that take into account the characteristics of wearable sensor systems has not been studied yet, leaving a gap in the research, which we believe has significant and novel consequences. Because in addition to the traditional notion of transferability - between different models trained on the same or disjoint subsets of a dataset - we also need to consider new dimensions when exploring adversarial transferability in wearable sensor systems. 

Wearable sensor systems are highly dynamic and have many different properties associated with it, which can impact the way these systems operate. For example, consider the placement of a wearable device on the human body. For human activity recognition, there are many devices available in the market today that can be worn on the body in various ways. Some are worn as a watch; others can be clipped on to clothes or shoes, strapped around the chest, and so forth. Furthermore, depending on the body location of the device, the sensor readings are very different, and consequently, machine learning algorithms trained on these sensor data learn unique mappings between inputs and outputs. Therefore an adversary who is planning to attack these types of machine learning systems must also take into account the different variations associated with a sensor system. These variations are well discussed in the literature \cite{okita, zeng} for the case of building inference models. But to the best of our knowledge, there has not been any work that had discussed these properties of wearable sensor systems from an adversarial point of view. All these lead to the fact that adversarial transferability in sensor systems is not simple and straightforward and has many nuances. We believe an extensive study of the adversarial transferability will not only show the strength of adversarial attacks but also mark its shortcomings and help us understand this unexplored problem space.

\subsection{Contributions}
In this paper, we present the first comprehensive evaluation of the transferability of adversarial examples in the context of wearable sensor systems. We not only consider the traditional notion of transferability but extend that with novel transferability directions, which we believe are unique to wearable sensor systems. In particular, we discuss the transferability of adversarial examples from the following four perspectives:
\begin{itemize}
    \item Adversarial transferability between machine learning models
    \item Adversarial transferability across subjects
    \item Adversarial transferability across sensor body locations
    \item Adversarial transferability between datasets.
\end{itemize}

And in doing so, in this work we make the following contributions:
\begin{enumerate}
    \item We, for the first time, introduce and define different types of adversarial transferability in the context of wearable sensor systems.
    
    \item We conduct extensive sets of experiments that highlight vulnerabilities and strengths of wearable sensor systems under different transferability cases for both targeted and untargeted evasion attacks.
    
    \item We discuss and validate our results with theoretical and graphical interpretations that take into account the properties of both models and data distribution.
    
    \item We discuss open problems and possible research directions for adversarial transferability in general for sensor systems.
\end{enumerate}

In the next section, we briefly discuss the human activity recognition pipeline and adversarial machine learning. We review the different properties of adversarial machine learning and describe several methods for generating adversarial examples. In section 3, we discuss our approach and explain the significance and consequence of our work. We explain why adversarial transferability in sensor systems have more facets than which is already known and discuss them in detail. We also describe our threat model in detail and introduce and define metrics used to measure adversarial transferability for both targeted and untargeted attacks. In the next section, we describe our experiments and results, which also include details about the hyper-parameters used for training different machine learning models. Finally, we discuss our results using ideas such as manifold learning and useful non-robust features and present our conclusion and recommendations along with a brief discussion on open problems for future research.

\section{Background}
Our discussion of the transferability of adversarial examples in the context of sensor systems uses human activity recognition as an example for experimentation purposes. Therefore, before we discuss adversarial machine learning and transferability in detail, we will first briefly explain the human activity recognition pipeline.

\subsection{Human Activity Recognition Pipeline}
The problem of human activity recognition can be defined as: Given a set $W = \{W_0,...,W_{m-1}\}$ of $m$ equally sized temporal window of sensor readings, such that each window $W_i$ contains a set of sensor reading $S = \{S_{i, 0},...,S_{i, k-1}\}$, and a set $A = \{a_0,...a_n-1\}$ of $n$ activity labels, the goal is to find a mapping function $f:S_i \to A$ that can be evaluated for all possible values of $S_i$ \cite{lara_survey_2013}. The activity recognition system generally consists of sensing, signal processing, signal segmentation, feature extraction and selection, and classification stages. Raw data from various sensors, such as accelerometer, gyroscope, and magnetometer, are collected and passed into the signal processing stage, where filtering and noise removal are applied to the sensor signals. The next stage is segmentation, where a continuous stream of the sensor values is divided into temporal windows. There are three types of segmentation methods: (1) activity-defined window; (2) event-defined window; and (3) sliding window. The sliding window method is used most widely due to its simplicity and real-time performance. After segmentation, statistical and structural features are extracted from each window segment and are used to train machine learning algorithms for activity classification. Another very successful approach to human activity classification uses convolutional neural network (CNN) with raw sensor segments as inputs. With CNN we don't need to compute features since the CNN model learns the features and the classifier simultaneously during the training process.

\subsection{Adversarial Machine Learning}
Given a machine learning classifier $f_\theta(x)$ characterized by the parameters $\theta$ and trained on dataset $D$ such that $D = \{(x, y)\}$, an adversary tries to find inputs that are formed by applying small but intentional perturbations $(\delta)$ to the original samples $x$ such that the perturbed inputs $\bar{x}$ are almost indistinguishable from the original samples and result in the classifier predicting an incorrect label $\bar{y}$ with high confidence. These perturbed input samples are called \emph{adversarial examples}. Hence, the objective of adversarial learning is to find perturbation $\delta$ which when added to the original inputs $x$ i.e., $\bar{x} = x + \delta$ changes the output of the classifier $f_\theta(\bar{x}) \neq y$. In general, an adversary can attack a machine learning system in three ways.

\begin{enumerate}
    \item \textit{Poisoning Attacks}: In poisoning attacks or confidence reduction, an adversary attempts to degrade the performance of a machine learning classifier by injecting adversarial examples during the training process to force the classifier to learn false connections between input and outputs. 

    \item \textit{Evasion Attacks}: Evasion attack is the most common type of adversarial attack carried out during test time. In general, an evasion attack involves getting a trained model to make mistakes on input samples. Evasion attack is sub-divided into two types: 1) untargeted attack, and 2) targeted attack. For an untargeted adversarial example $\bar{x}$, an adversary intends to missclassify $\bar{x}$ into any class other than its true class i.e., $f_\theta(\bar{x}) \neq y$ such that $f_\theta(x) = y$. For a targeted adversarial example $\bar{x}$, the adversary defines the target class $\bar{y}$ in which it wants to have the classifier classify the adversarial example i.e., $f_\theta(\bar{x}) = \bar{y}$.
    
    \item \textit{Exploratory Attacks}: In exploratory attacks, the adversary tries to gain as much knowledge as possible about the learning algorithm of the target system and patterns in the training data. 
\end{enumerate}

In this work, we only consider evasion attack methods because we discuss adversarial transferability at inference time. And, the difficulty in mounting evasion attacks against a target system is heavily influenced by the knowledge an adversary has about the target system. The extent of an adversary's knowledge about the target system dictates the setting in which it operates.

\begin{enumerate}
    \item \textit{White-box Setting}: A white-box setting assumes that the adversary has complete knowledge about the target system. It includes anything related to the target system such as dataset, model architectures, hyper-parameters values, activation functions, number of layers, and model weight. This comprehensive knowledge about the target system makes it easier for the adversary to mount successful evasion attacks. In this mode, the adversary can compute adversarial examples using the target system and don't have to rely on the transferability property of adversarial examples.
    
    \item \textit{Black-box Setting}: In the black-box setting, the adversary has no knowledge of the target system. The adversary only has access to an oracle to the target system to submit inputs and observe outputs. Evasion attacks in a black-box setting exploit the transferability properties of adversarial examples to fool the target system.
\end{enumerate}

The difficulty of operating in a black-box setting is mitigated by exploiting the transferability property of adversarial examples. It has been shown that adversarial examples computed for one machine learning system are also effective against other machine learning system \cite{Papernot16}. For example, adversarial examples computed for a DNN model are also highly successful against support-vector machines (SVM). Adversarial transferability is not only valid for models operating in the same domain, but also across domains \cite{naseer2019cross}. In our earlier discussion of the operating model, Bob, an adversary, operates in the white-box setting with respect to source system $S_1$ and black-box setting with respect to the target system $S_2$. Hence, Bob depends on the transferability property of adversarial examples computed using $S_1$ to fool the target system $S_2$. 


\subsection{Methods of Generating Adversarial Examples}
Now that we have some understanding about different types of adversarial attacks and how the knowledge of an adversary defines the context in which it operates. We shift our attention to different types of evasion attacks methods we have used in our experiments. The fundamental condition when computing adversarial example is that the perturbation $\delta = \{\delta_1, \delta_2, \dots , \delta_n\}$ added to the benign samples $x = \{x_1, x_2, \dots, x_n\}$ to get adversarial samples cannot be large. This requirement is satisfied bu bounding the adversarial perturbation $\delta$ with some adversarial budget $\epsilon$ using $l_p-$norms, where $p \in \{0, 1, 2, \infty\}$. For a model $f_\theta$, adversarial examples $\bar{x}$ are defined as the solution to the following optimization problem.

\begin{equation}
    \bar{x} = x + \argmin_\delta\{\|\delta\|: f(x + \delta) \neq f(x)\}
    \label{equ:adv_min}
\end{equation}

Here, $\|.\|$ is a type of $l_p-$norm depending on the method used to compute adversarial examples. For two vectors $x$ and $\bar{x}$, $l_0$ counts the number of elements in $\bar{x}$ that has changed its values compared to $x$, $l_2$ measure the Euclidean distance between the two vectors, and $l_\infty$ denotes the maximum changes for all elements in the vector $\bar{x}$. The attack methods we discuss in the next section, solve the above optimization problem \ref{equ:adv_min} in one way or another.

\subsubsection{Fast Gradient Sign Method (FGSM)}
Fast gradient sign method (FGSM) was proposed by \textit{Goodfellow et al.,} \cite{GoodFellow14} and is one of the simplest and computationally efficient method to compute adversarial examples. FGSM computes the adversarial perturbation by calculating the gradient of the loss function of the model with respect to the input. This method solves the following optimization problem to maximize the loss such that adversarial perturbations are bounded by $\epsilon$ subject to $l_\infty$ norm. 

\begin{equation}
    \bar{x} = x + \epsilon * \textrm{sign}(\nabla_xJ_\theta(x, y))
\end{equation}

Here, $J_\theta$ is the loss function, and $\nabla_x$ denotes the gradient of the loss function with respect to the input $x$, and $y$ is the actual label \cite{Chakraborty_2018}. For targeted examples, FGSM minimizes the loss function with respect to input $x$ such that the modified input is classified into the target class $\bar{y}$ specified by the adversary.

\begin{equation}
    \bar{x} = x - \epsilon * \textrm{sign}(\nabla_xJ_\theta(x, \bar{y}))
    \label{equ:fgsm_tar}
\end{equation}

Notice the change in sign and also the presence of the target class $\bar{y}$ in the optimization equation \ref{equ:fgsm_tar}. For targeted case, we are trying to find adversarial perturbations $\delta$ that decrease the loss of the model for the target class $\bar{y}$ and for untargeted case we find adversarial perturbations the increase the loss of the model in general.

\subsubsection{Basic Iterative Method (BIM)}
Basic iterative method (BIM) is a straightforward extension to the fast gradient sign method and runs FGSM $n$ number of times with a small step size until some requirements are satisfied. Iteratively running FGSM allows the adversary to search the model input space more thoroughly to find optimal perturbations.

\begin{equation}
    \begin{split}
        \bar{x}_0 & = x \\
        \bar{x}_{n+1} & = \textrm{Clip}_{x, \epsilon} \{\bar{x}_n + \alpha * \textrm{sign}(\nabla_x J_\theta(x, y))\}
    \end{split}
    \label{equ:biter}
\end{equation}

Here, $\alpha$ is the step size and is usually given by $\alpha = \epsilon / n$. $\textrm{Clip}_{x, \epsilon}(A)$ denotes the element-wise clipping of $A$, such that the range of $A_{i,j}$ after clipping is in the interval $[x_{i, j} - \epsilon, x_{i, j} + \epsilon]$. The basic iterative method can also be used to compute targeted adversarial examples by the simple modification of sign reversal and the introduction of the target class in the optimization equation \ref{equ:biter}.

\subsubsection{Jacobian-Based Saliency Map Attack (SMM)}
Jacobian-based saliency map attack \cite{PapernotMJFCS15} first computes the Jacobian matrix of the given input $x$,
\begin{equation}
    J_F(x) = \frac{\partial F(x)}{\partial x} = \left[ \frac{\partial F_j(x}{\partial x_i}\right]_{i x j}
\end{equation}

Here, $F$ is the second-to-last layer logits of the neural network. Jacobian-based saliency map attack finds input features of $x$ that cause the most significant changes to the output of the model and computes perturbations that induce significant output variations so that a change in a small portion of features of $x$ could fool the neural network \cite{adv_survey_yuan}.

\subsubsection{Carlini-Wagner Attack (CW)}
The Carlini-Wagner attack was proposed to defeat defensive distillation \cite{carlini2017towards}, and solves the following optimization problem to find adversarial perturbations.

\begin{equation}
    \begin{split}
        & \min \|\delta\|_p \\ 
        \textrm{subject to} \quad & C(x + \delta) = t, \quad x + \delta \in [0, 1]^n    
    \end{split}
\end{equation}
where $C(x)$ is the class label returned for input $x$ and the noise level is measured using either $l_0, l_2$ or $l_\infty$ norm. Carlini-Wagner attack is considered one of the best evasion attack method and computes adversarial examples by finding the smallest noise $\delta \in R^{nxn}$ which changes the classification of the model to a class $t$.

\subsubsection{Momentum Iterative Attack (MIM)}
Momentum iterative attack method \cite{dong2018boosting} integrates the concept of momentum into the basic iterative method to generate adversarial examples for targeted and untargeted cases using $l_2$ and $l_\infty$ norms respectively. The momentum is a technique for accelerating gradient descent algorithms by accumulating a velocity vector in the gradient direction of the loss function across iterations \cite{dong2018boosting}. The introduction of momentum helps the method achieve optimum results faster by stabilizing update directions and escaping from poor local maxima. 

\section{Approach}
In this section, we discuss the approach we have used to evaluate adversarial transferability in wearable sensor systems. We highlight and explain different types of adversarial transferability that we believe are inherent to wearable sensor systems and present the threat model used in our experiments. We also define metrics used to measure untargeted and targeted adversarial transferability. Finally, we discuss datasets used in our experiments and outline some conditions that are crucial for the sound analysis of adversarial transferability in wearable sensor systems.

\subsection{Adversarial Transferability}
Human activity classification usually employs some wearable device such as a smartwatch, smartphone, smart shoes, chest band, fitness band to detect and measure physical activities. The underlying machine learning algorithms use the data from sensors to learn the characteristics of different activities. But depending on the properties of the sensors, the sensor reading can vary significantly even though the sensors are trying to measure and detect the same physical phenomena. This is because human activities are highly complex, dynamic, and diverse, and the sensor readings for an activity vary significantly even if the same person performs the same activity under similar conditions compared to say image classification where an image of a dog is always a dog independent of the presentation and context. These variations in the sensor reading eventually reach machine learning algorithms making them learn unique mappings from inputs to outputs and creating newer challenges and opportunities for an adversary that wants to attack these systems. From differences in the electrical properties of the sensor to the location of wearable devices on the human body, there are numerous ways in which different aspects of the wearable sensor systems can affect adversarial transferability. Therefore adversarial transferability in the context of wearable sensor systems has many newer dimensions than the traditional notion of transferability and requires a detailed study to evaluate and understand its cause and consequences. To this end in this work, we study adversarial transferability from the following four perspectives:

\begin{itemize}
    \item \textbf{Adversarial transferability between machine learning models}\newline
    The transferability between different machine learning algorithms trained on the whole or subset of the same dataset is the default and the most discussed variety of adversarial transferability. To exploit this mode of transferability, the adversary computes adversarial examples using one machine learning model and then performs adversarial attacks on other models using the generated adversarial examples. 
    
    \item \textbf{Adversarial transferability across subjects} \newline
    In sensor systems, the dataset used to train machine learning algorithms is collected using human subjects. This is similar to an image dataset where images of various objects are captured using different types of cameras. But what separates human subjects from the optical sensor in cameras is that human subjects inject biases in the data that are personalized to each individual and are hard to eliminate with preprocessing. Having artifacts associated with individuals gives the problem of adversarial transferability in wearable sensor systems a new direction. The adversary can leverage the biases injected by individuals to design better attack methods or suffer from this when trying to attack a target system. Hence, evaluating adversarial transferability between machine learning systems trained on data from different subjects becomes very important.

    \item \textbf{Adversarial transferability across sensor body locations} \newline
    Another essential attribute present in wearable sensor systems is the body location of the sensor system. For example, activity trackers can be worn in many different ways. Some can be worn as a wristwatch or band, others can be clipped on to clothes and shoes or placed inside pockets, and some can be even worn as jewelry. For two sensors of the same type - one wrapped around the subject's chest and other worn on the wrist - the sensor readings depend heavily on the orientation of the sensor and other factors. And these differences in the sensor readings affect the mapping learned by machine learning algorithms between inputs and outputs and consequently transferability of adversarial examples.

    \item \textbf{Adversarial transferability between datasets} \newline 
    The final and most complex type of adversarial transferability is transferability between machine learning systems - same or of different architectures - trained on different datasets. For example, for the task of human activity recognition, different manufacturers use different types of sensors and collect proprietary datasets to train machine learning algorithms. Now for an adversary which has access to say, system $S_1$, and its dataset $D_1$, it is challenging to make the adversarial examples computed using $M_1$ transferable to some other system $S_2$ trained on dataset $D_2$. The challenges can stem from bias from subjects, sensors placed at different sensor positions, different types of sensors, and different data processing techniques. Hence, in the final part of our experiments, we explore adversarial transferability from this direction.
\end{itemize}

\subsection{Threat Model}
\begin{figure}[tbh!]
    \centering
    \includegraphics[width=0.8\linewidth]{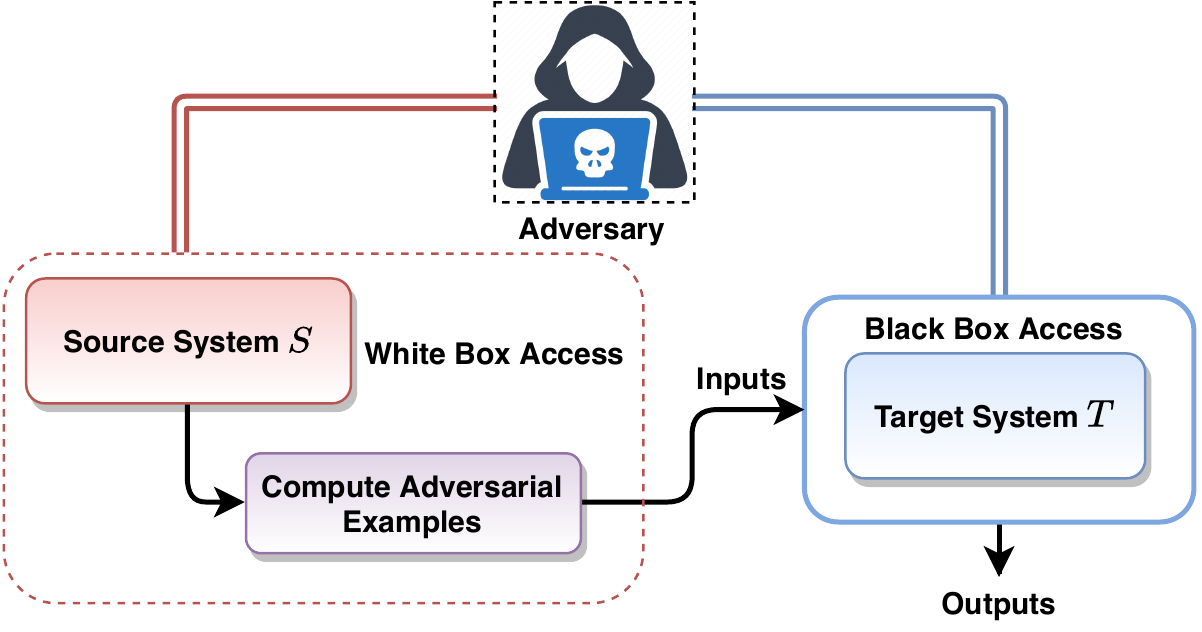}
    \caption{Threat model in which an adversary operates with complete access to the source system $S$ and dataset $D_S$ but only oracle access to the target system $T$. The adversary has no knowledge about the target system and its properties, such as dataset used for training, type of learning algorithm, and hyper-parameters of the model. The adversary can only compute adversarial examples using the source model $S_m$ and transfer the adversarial examples to target system $T$ in the hope of fooling target model $T_m$ by exploiting the transferability properties of adversarial examples. This assumption allows us to study the transferability of adversarial examples, which would not have been possible if the adversary can compute adversarial examples using the target system $T$. Our assumption draws merits from the fact that in practical situations, the adversary rarely has complete access to the target system, and the adversary needs to learn and guess details about the target system to attack it.}
    \label{fig:threat_model}
    \Description[Threat Model]{}
\end{figure}

Depending on the type of adversarial transferability the adversary wants to exploit to attack the target system, the adversary operates in different settings. In general, the threat model has two main components. The first part concerns the target system $T$, which the adversary wants to attack and the second part takes into account the source system $S$ which contains the source model $S_m$. The adversary can only send inputs to the target system and observe the class prediction, and hence operates in the black-box setting with respect to the target system $T$. The adversary has white-box access to the source system $S$ and can compute adversarial examples using the source model $S_m$. Here the objective of the adversary is to fool the target system $T$ on the adversarial examples computed using the source model $S_m$. Fig \ref{fig:threat_model} shows the graphical representation of the threat model. In all four cases of adversarial transferability discussed in the next sections, the adversary computes adversarial examples using the source model $S_m$ trained on source dataset $D_S$ and attacks the target system $T$ with machine learning algorithm trained on target dataset $D_T$. For transferability between machine learning models $D_S$ and $D_T$ are the same and for the remaining cases of adversarial transferability datasets $D_S$ and $D_T$ are different. In adversarial transferability across subjects, the source dataset contains samples from a group of subjects and the target dataset contains samples from the remaining subjects. All other characteristics of sensor and data processing stages remains the same for source and target dataset. In adversarial transferability across sensor body locations, the source dataset contains sensor reading from sensor placed at one body position, for example right-wrist, and the target dataset contains sensor values from sensor placed at another body position, say chest. Finally, in adversarial transferability between datasets, the source and target dataset are completely different and have different distributions.

\subsection{Measuring Adversarial Transferability}
In the simplest sense, when we measure adversarial transferability, we are saying how many of the adversarial samples designed for the source system can fool the target system. Usually, the metric used to measure the performance of a machine learning is accuracy, defined as the ratio of the number of samples correctly classified to the total number of samples. But measuring accuracy for untargeted attacks is counter-intuitive in the sense that, for untargeted attacks, the mistakes made by the target system on adversarial examples tell us more about the effectiveness of the attack than the accuracy. Hence, to measure adversarial transferability in both untargeted and targeted case, we introduce a new metric called \textit{Success Score} $(SC)$. The success score defined in percentage is the ratio of the number of adversarial examples that were able to fool the target system $(N_t)$ to the total number of samples $(N)$
\begin{equation}
    \textrm{Success Score (SR)} = \frac{N_t}{N} * 100
\end{equation}
For untargeted case, $N_t$ is equal to the number of adversarial examples that were classified into any class other than their true class by the target system and for targeted case, $N_t$ equals the number of adversarial examples that were classified into the target class by the target system.



\subsection{Datasets}
In our experiments, we have used three real-world datasets to evaluate different types of adversarial transferability in the context of sensor systems. Since we are taking human activity recognition as an example system, we have conducted our experiments with 3-axial accelerometer data. 

\begin{enumerate}
    \item \textit{UCI dataset}\footnote{\url{https://archive.ics.uci.edu/ml/datasets/human+activity+recognition+using+smartphones}} \cite{Anguita13} was compiled from a group of $30$ participants, each wearing a smartphone on their waist and performing $6$ different activities: \textit{standing, sitting, laying down, walking, walking upstairs} and \textit{walking downstairs} in a lab setting. Data from $3$-axial accelerometer and gyroscope sensors were sampled at a frequency of $50$ Hz and pre-processed by applying noise filters.
    

    \item \textit{MHEALTH dataset}\footnote{\url{https://archive.ics.uci.edu/ml/datasets/MHEALTH+Dataset}} \cite{banos2015design} consists of body motion and vital signs recording of $10$ volunteers of different profiles while performing $12$ different physical activities in an out-of-lab environment without any constraint. \textit{Shimmer2} wearable device placed on the subject's chest, right wrist, and left ankle were used to measure the motion experienced by the diverse body parts using a 3-axial accelerometer at a frequency of $50$ Hz. The class \textit{Jump Front \& Back} has fewer number of samples compared to other classes. Therefore, to balance the dataset, we have removed the samples from the \textit{Jump Front \& Back} class from our analysis. 
    
    
    
    \item Daily Log (DL) dataset \footnote{\url{https://sensor.informatik.uni-mannheim.de/\#dataset_dailylog}} \cite{sztyler2015discovery} has accelerometer, orientation, and GPS sensor data collected from $7$ individuals using a smart-phone and smart-watch with a self-developed sensor data collector and labeling framework. Acceleration and orientation sensors were sampled at $50$ Hz and GPS data was collected every $10$ minutes. The data was collected when participants were doing their daily routine and it was up the participants were the device should be positioned on the body. We randomly select subset of the data to use in our experiments such that each activity class has the same number of samples.
    
    \begin{figure}[tbh!]
      \centering
      \includegraphics[width=0.8\linewidth]{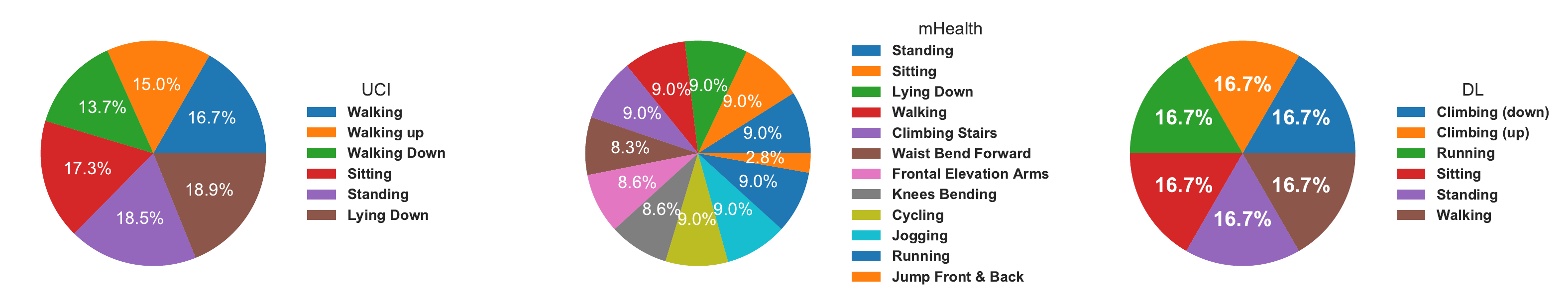}
      \caption{Activity distribution of the three datasets (best viewed in color)}
      \label{fig:dataset_pie}
      \Description[Activity distribution for the three datasets in pie chart form.]{This plot shows the percentage of each activity class in the total dataset for the UCI, mHealth, and DL dataset.}
    \end{figure}
    
\end{enumerate}
\begin{table}[tbh]
    \caption{Characteristics of the three datasets used in the analysis ($\#$ denotes ``number of'').}
    \label{tab:datasets}
    \begin{tabular}{c c c c c c c}
        \toprule
        {Dataset} & {$\#$ Subject} & {$\#$ Activities} & {Frequency} & {Window Size (Seconds)} & {$\#$ Devices} & {$\#$ Samples}\\
        \midrule
        UCI & 30 & 6 & 50 Hz & 2.56 & 1 & 10299 \\ 
        MHEALTH & 10 & 12 & 50 Hz & 2.56 &  3 & 5133\\
        DL & 7 & 6 & 50 hz & 2.56 & 2 & 16434 \\
        \bottomrule
    \end{tabular}
\end{table}

Now that we understand the different characteristics of the three datasets, we need to establish some conditions so that the analysis of different types adversarial transferability across the datasets is possible and sound.
\begin{enumerate}
    \item \textbf{Sampling Frequency:} One of the criteria that we have used to select datasets for our experiments is the sampling frequency of the sensor. For all real-world datasets used in this paper, the sampling frequency is $50$ Hz, which is considered adequate for human activity \cite{1612909}.
    
    \item \textbf{Input Size:} The length of the window segment in all datasets must be the same because we cannot train machine learning algorithms with variable input sizes. In our experiments, we have set the length of the raw sensor segment to $128$, which corresponds to the window size of $2.56$ seconds at sampling frequency of $50$ Hz. Setting the window size to $128$ was motivated by the fact that a window size of $1-2$ seconds with $50\%$ overlap is considered a good choice for activity classification \cite{Banos14}. 
    
    \item \textbf{Data Scaling:} The range of values in the three datasets are very different. 
    Hence, it becomes vital that we standardize the datasets such that the analysis across different datasets is possible. For this we use the \textit{MinMaxScaler} with range set to [-1.0, 1.0] from the \emph{sklearn} library \cite{scikit-learn} to scale the dataset. \textit{MinMaxScaler} is the least disruptive to the information in the original data and preserves the shape of the data and does not reduce the importance of outliers.
    
    \item \textbf{Activity Classes:} Another important factor when choosing the dataset for our experiments was the activity classes. The baseline condition was that there should be some activity classes that are common for all datasets because this would allow us to analyze targeted adversarial transferability between datasets. The activities \textit{walking, sitting, standing} and \textit{climbing stairs (walking up)} are common to all three datasets. Also, having activities classes that are not common between the datasets further helps us analyze transferability with generalization.
\end{enumerate}

\section{Experimental Results}
In this section, we discuss our experiments and results. We discuss four cases of adversarial transferability, and for each case, we have analysis for both targeted and untargeted evasion attacks. 

\label{section:adv_tr_bet_models}
\subsection{Adversarial Transferability Between Machine Learning Models}
To analyze the adversarial transferability between machine learning models, we train different six different machine learning algorithms for a common dataset and compute adversarial examples using one of the trained model. We use the feature dataset of all three real-world datasets for training the machine learning algorithms. Using the feature data enables us to train different kinds of machine learning algorithms, which is not possible using the raw sensor data. We compute $45$ statistical features commonly used in HAR \cite{zhang2011feature}, from the sensor segments of all three datasets. We have selected the following algorithms to analyze adversarial transferability between machine learning models: 1) Support Vector Classifier (SVC), 2) Random Forest Classifier (RFC), 3) K-Nearest Neighbor Classifier (KNN), 4) Decision Tree Classifier (DTC), 5) Logistic Regression Classifier (LRC), and 6) Deep Neural Network (DNN). Table \ref{tab:clfs_accuracies} shows the classification accuracy of all trained models on the training and test set for the three datasets. In general, all trained models have very high classification accuracy on training and test sets. The deep neural network (DNN) has $3$ layers with $64$, and $32$ neurons in the first and second layers. In the last layer the number of neurons is equal to the number of activity classes of the respective datasets. $l2$-regularization with coefficient $0.001$ and ReLU activation is used in the first and second layers, and the output layer has Softmax activation. \textit{TensorFlow} \cite{tensorflow2015} was used to train the deep neural network with hyper-parameters: $200$ epoch, mini-batch size of $32$, Adam \cite{kingma2014adam} optimizer with learning rate $0.001$, and sparse categorical cross-entropy loss. All other classifiers were trained using the sklearn library \cite{scikit-learn}. The maximum iteration for SVC was set to $5000$ with scaled gamma, and the number of estimators for RFC was set to $100$. For logistic regression, the LBFGS solver was used with $5000$ maximum iterations and for K-Nearest Neighbors the number of components was set to $5$. All other parameters of classifiers were left to their default values.  

\begin{table}[tbh]
    \caption{The classification accuracy of different machine learning algorithms on the training and test set of all three datasets.}
    \label{tab:clfs_accuracies}
    \begin{tabular}{c}
        \toprule
        \multicolumn{1}{c}{ML} \\
        {Algorithms}\\
        \midrule
        SVC \\
        RFC \\
        KNN \\
        DTC \\
        LRC \\
        DNN \\
        \bottomrule
    \end{tabular}
    \quad
    \begin{tabular}{cc}
        \toprule
        \multicolumn{2}{c}{UCI} \\
        \cline{1-2} 
        {Train Set} & {Test Set}\\
        \midrule
        76.20\% & 76.38\% \\
        100.0\% & 84.85\% \\
        84.85\% & 79.10\% \\
        100.0\% & 72.93\% \\
        75.49\% & 76.54\% \\
        84.90\% & 82.05\% \\
        \bottomrule
    \end{tabular}
    \quad
    \begin{tabular}{cc}
        \toprule
        \multicolumn{2}{c}{MHEALTH} \\
        \cline{1-2} 
        {Train Set} & {Test Set}\\
        \midrule
        90.40\% & 90.46\% \\
        100.0\% & 96.39\% \\
        97.56\% & 96.07\% \\
        100.0\% & 92.22\% \\
        91.15\% & 89.90\% \\
        99.25\% & 97.19\% \\
        \bottomrule
    \end{tabular}
    \quad
    \begin{tabular}{cc}
        \toprule
        \multicolumn{2}{c}{DL} \\
        \cline{1-2} 
        {Train Set} & {Test Set}\\
        \midrule
        87.79\% & 87.61\% \\
        100.0\% & 89.87\% \\
        91.96\% & 87.90\% \\
        100.0\% & 85.08\% \\
        86.12\% & 85.66\% \\
        94.75\% & 89.82\% \\
        \bottomrule
    \end{tabular}
\end{table}


To evaluate these classifiers for adversarial transferability between machine learning models, we compute targeted and untargeted adversarial examples using the DNN model - the source model. We choose to use the DNN model to compute adversarial examples, because evasion attacks methods based on gradient optimization are more mature and there are large number of successful attack methods available for neural networks \cite{GoodFellow14, carlini2017towards}. There are some adversarial attack methods that can compute adversarial examples with non-parametric models such as decision trees and k-nearest neighbors \cite{Papernot16, yang2020robustness} and we explore these attacks later in our discussion section. We have used five different adversarial attack methods to compute adversarial examples for both targeted and untargeted attacks: 1) Fast Gradient Sign Method (FGSM) \cite{GoodFellow14}, 2) Basic Iterative Method (BIM), 3) Saliency Map Method \cite{GoodFellow14}, 4) Carlini-Wagner Method (CW), and 5) Momentum Iterative Method (MIM) \cite{carlini2017towards}. These attack methods differ in their optimization approaches and complexity. The \textit{CleverHans} \cite{papernot2018cleverhans} library was used to compute the adversarial examples with the following parameters:
\begin{enumerate}
    \item Since the range of feature data is $[-1.0, 1.0]$, we set the adversarial perturbation budget to $\epsilon = 0.5$.
    \item For clipping the range is set to $[-1.0, 1.0]$.
    \item The number of iterations where for basic iterative method and momentum iterative method is set to $50$.
    \item The value of perturbation per iteration is $0.5/50 = 0.01$.
\end{enumerate}

\begin{figure}[tbh!]
    \centering
    \includegraphics[width=0.9\linewidth]{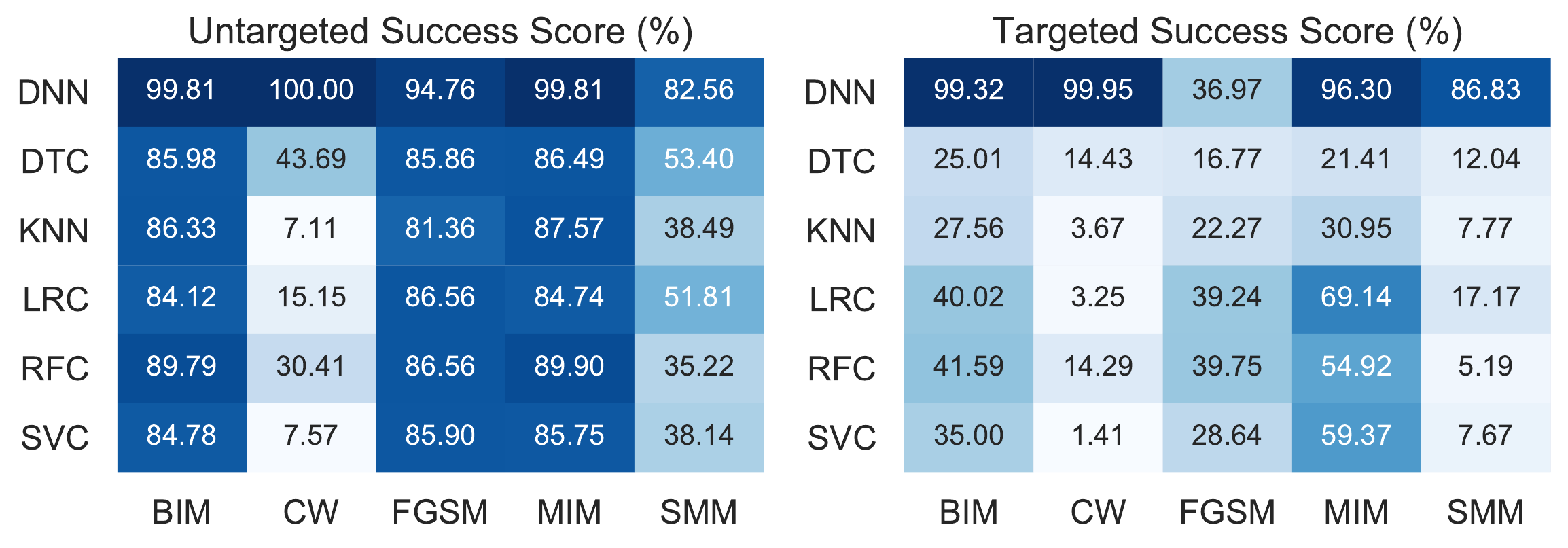}
    \caption{Success score of untargeted and targeted adversarial examples for the UCI dataset.}
    \label{fig:uci_trans_model}
\end{figure}

\begin{figure}[tbh!]
    \centering
    \includegraphics[width=0.9\linewidth]{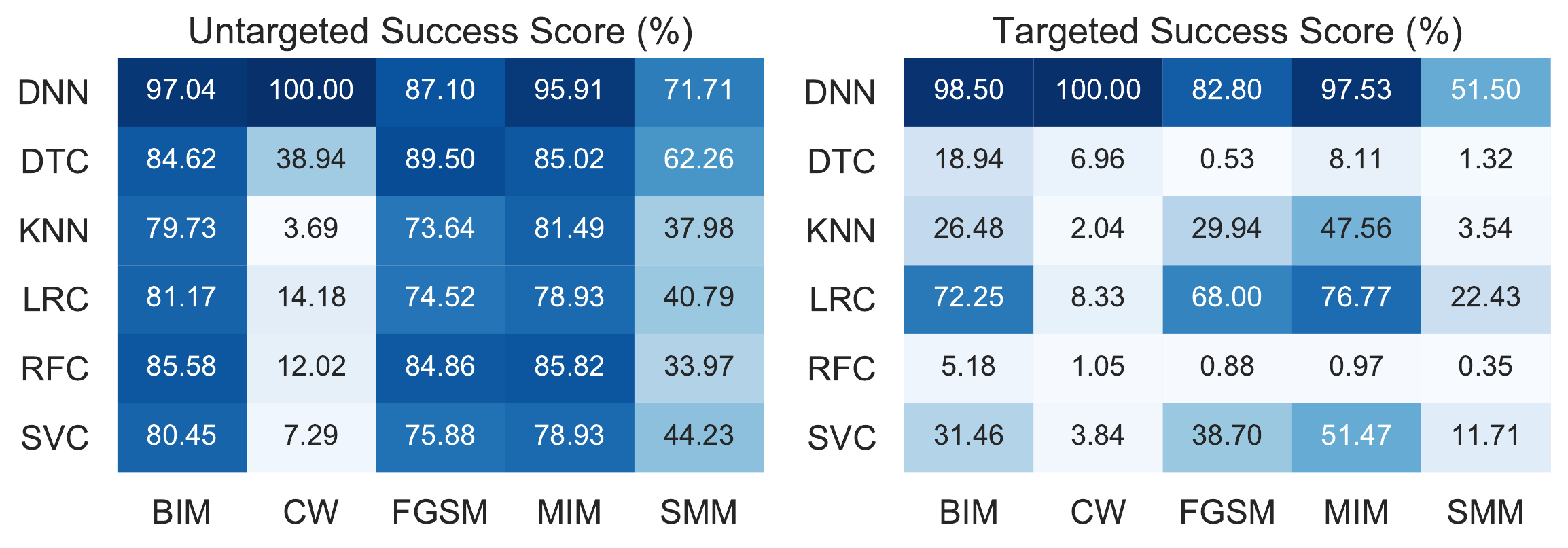}
    \caption{Success score of untargeted and targeted adversarial examples for the MHEALTH dataset.}
    \label{fig:mh_trans_model}
\end{figure}

\begin{figure}[tbh!]
    \centering
    \includegraphics[width=0.9\linewidth]{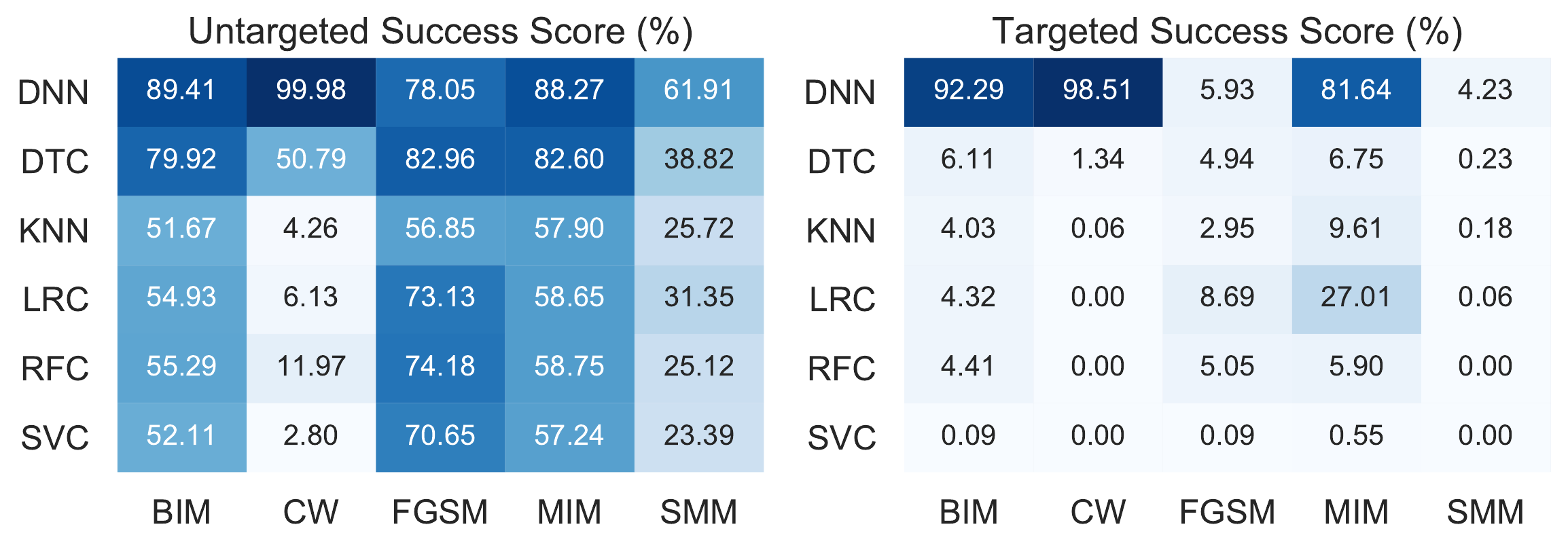}
    \caption{Success score of untargeted and targeted adversarial examples for the DL dataset.}
    \label{fig:dl_trans_model}
\end{figure}

For both untargeted and targeted attack, the adversary has complete access to the DNN system and the common dataset used to train all classifiers. For the targeted attack we selected the ``\textit{Sitting}'' activity class as the target class because it is common across all three datasets. Figures \ref{fig:uci_trans_model}, \ref{fig:mh_trans_model}, and \ref{fig:dl_trans_model} shows the success score of untargeted and targeted adversarial examples for the UCI, MHEALTH, and DL datasets. Different adversarial attack methods, shown along the columns, were used to generate adversarial examples using the DNN model, and generated adversarial examples were used to attack different machine learning models, shown along the rows. Each number in the heatmap, shows the success score of untargeted/targeted adversarial examples computed using the DNN model for the attack method specified by the column index on the machine learning model denoted by the row index. For example, in figure \ref{fig:uci_trans_model} the success score of untargeted adversarial examples computed with adversarial attack BIM on the SVC model is $84.78\%$ and the success score of targeted adversarial examples for the target class ``\textit{Sitting}'' is $35\%$. Likewise for DL dataset, figure \ref{fig:dl_trans_model}, the success score for the same combination of adversarial attack method and machine learning model is $52.11\%$ for untargeted adversarial examples and $0.09\%$ for targeted adversarial examples.

In general, we found high adversarial transferability between machine learning models for untargeted adversarial examples. For targeted attacks, adversarial examples were less transferable for all three datasets. In particular, we found Decision Tree Classifier (DTC), K-Nearest Neighbor (KNN), and Random Forest Classifier (RFC) to be more robust towards targeted adversarial examples computed using the Deep Neural Network (DNN) model. We also found that the level of adversarial transferability between machine learning systems differ greatly across the three datasets. For the DL dataset, both targeted and untargeted adversarial examples were less likely to be transferable. For targeted attacks, the adversarial transferability was very poor with success score values of $0.0\%$ in many cases. We believe the lower success score of targeted adversarial examples in general is due to fundamental differences between the targeted and untargeted attacks. An untargeted attacks is considered successful if an input is classified into any class other than its actual class by the target system but for the targeted attack to be successful the input must be classified into the target class by the target system. Hence, targeted attack are much more difficult and an adversary will have higher success score with untargeted attack compared to targeted attack for the same level of adversarial perturbation. Also, we suspect the lower success score for the DL dataset is due to nature of the dataset. The DL dataset was collected in daily-living conditions while participants were following their daily routine compared to the MHEALTH and UCI dataset which were collected in a lab environment. Collecting sensor data in daily-living conditions can induces noise and artifacts in the sensor data and as a result different learning algorithms will learn different mappings from input to output. Furthermore, no data preprocessing is applied to the DL dataset but both UCI and MHEALTH dataset undergo noise removal and filtering. Consequently, adversarial transferability which aims to capitalize on the common input-output mappings shared by different machine learning algorithms to fool a target system on adversarial examples computed using the source system will not be very successful.

\subsection{Adversarial Transferability Across Subjects}
In sensor systems, datasets used to train machine learning algorithms are usually collected from human subjects. Human subjects introduces individual biases and artifacts in the sensor readings that are characteristics of each user. For example, every individual has subtle difference in gait and body movements for common physical activities such as sitting, running, standing etc. These difference are encoded into sensor readings when sensor data are collected and in turn machine learning algorithms trained on the sensor data learns common mappings as well as unique connections characteristics of individual subjects. Hence, to better understand the effects of adversarial transferability in wearable sensor systems, we need to take this aspect of wearable systems into account. In this section, we study how the biases introduced by human subjects in the sensor readings affect the transferability of adversarial examples. We train two machine learning algorithms of same type and architecture and with same hyper-parameters on different datasets. The adversary has white-box access to the source model, and black-box access to the target system. The adversary computes adversarial examples using the source system and attacks the target system using the computed targeted and untargeted adversarial examples. All three datasets has subject ID associated with each row of sensor readings or data files. We randomly select data from half the subject to create the dataset for the source system, called source dataset, and the data from the remaining half subjects is used in the target system, called target dataset. The adversary has complete access to the source dataset and no access to the target dataset. Table \ref{tab:source_target_dataset_info} shows the numbers of samples in source and target datasets for UCI, MHEALTH, and DL datasets used in the analysis. We have decided to use $1-$D convolutional neural network (CNN) for both source and target system machine learning algorithm because of its simplicity and superior performance. Furthermore, CNN allows us to use the raw sensor data directly to train the model without needing to compute features from the sensor segments. The CNN model is trained using the Adam \cite{kingma2014adam} optimizer with a learning rate of $0.001$ and no learning rate decay. The loss of the model is computed using the categorical cross-entropy loss function and the architecture of the CNN model is as follows: 

\begin{itemize}
    \item The first layer or the input layer is a $1-$D CNN layer with $100$ filters, kernels of size $10$, strides of $2$, and ReLU activation.
    \item The second layer is a $1-$D CNN layer with $50$ filters, kernel size of $5$, strides of $1$, and ReLU activation. 
    \item The third layer is a $1-$D Global Max Pooling layer. 
    \item The fourth layer is a fully-connected layer with $64$ neurons and ReLU activation. We have set the drop out coefficient for this layer to $0.3$.
    \item The last layer or the output layer is a fully connected layer with number of neurons equal to the number of activity classes for each dataset. The activation function for this layer is the Softmax function. 
\end{itemize}

\begin{table}[tbh]
    \caption{Details about the source and target dataset fashioned by randomly selecting subjects data for the UCI, MHEALTH, and DL datasets. Here, $\#$ means ``number of''.}
    \label{tab:source_target_dataset_info}
    \begin{tabular}{ccccccc}
        \toprule
        {Dataset} & {$\#$ Samples} & {$\#$ Subjects} & {$\#$ Source Subjects} & {$\#$ Source Samples} & {$\#$ Target Subjects} & {$\#$ Target Samples}\\
        \midrule
        UCI & 10299 & 30 & 15 & 5138 & 15 & 5161 \\
        MHEALTH & 5133 & 10 & 5 & 2464 & 5 & 2527\\
        DL & 16434 & 7 & 3 & 9918 & 4 & 6516\\
        \bottomrule
    \end{tabular}
\end{table}

Table \ref{tab:source_target_clfs_acc} shows the classification performance of the source and target system of UCI, MHEALTH, and DL datasets on their respective source and target datasets. For the UCI and MHEALTH dataset, the source and target models have high classification accuracy on both source and target datasets. High classification accuracy of the source model on the target dataset and target model on the source model demonstrates high level of generalization for both source and target systems. Therefore, in theory adversarial examples that are effective against source system should also be effective against target system because the input-output mapping learned by the source system and target system share common knowledge and an adversary should be able to exploit these common mappings to fool the target system with the adversarial examples computed using the source system. On the other hand, the classification accuracy of the source system on the target dataset and the target system on the source dataset is low for the DL dataset. Low classification accuracy means that target and source system of the DL dataset, has less shared knowledge between them and consequently the adversarial transferability should be poor as well. When computing adversarial examples, the adversary can set different values for the perturbation budget ($\epsilon$). Since, all our datasets has a range of $[-1.0, 1.0]$, we have decided to use the following values of $\epsilon$ in our experiments $\epsilon \in \{0.1, 0.25, 0.5, 0.9\}$.

\begin{table}[tbh]
    \caption{The classification accuracy of different machine learning models on the training and test set of all three datasets.}
    \Description[The classification accuracy of different machine learning models on the training and test set of all three datasets.]{Same training set was used to train different machine learning algorithms and the trained models were evaluated on the same test set.}
    \label{tab:source_target_clfs_acc}
    \begin{tabular}{c}
        \toprule
        \multicolumn{1}{c}{Machine Learning} \\
        {System}\\
        \midrule
        Source \\
        Target \\
        \bottomrule
    \end{tabular}
    \quad
    \begin{tabular}{cc}
        \toprule
        \multicolumn{2}{c}{UCI} \\
        \cline{1-2} 
        {Source Set} & {Target Set}\\
        \midrule
        81.49\% & 61.42\% \\
        62.43\% & 85.58\% \\
        \bottomrule
    \end{tabular}
    \quad
    \begin{tabular}{cc}
        \toprule
        \multicolumn{2}{c}{MHEALTH} \\
        \cline{1-2} 
        {Source Set} & {Target Set}\\
        \midrule
        99.66\% & 66.55\% \\
        81.37\% & 99.72\% \\
        \bottomrule
    \end{tabular}
    \quad
    \begin{tabular}{cc}
        \toprule
        \multicolumn{2}{c}{DL} \\
        \cline{1-2} 
        {Source Set} & {Target Set}\\
        \midrule
        81.74\% & 25.28\% \\
        34.09\% & 86.05\% \\
        \bottomrule
    \end{tabular}
\end{table}

\begin{figure}[tbh!]
    \centering
    \includegraphics[width=\linewidth]{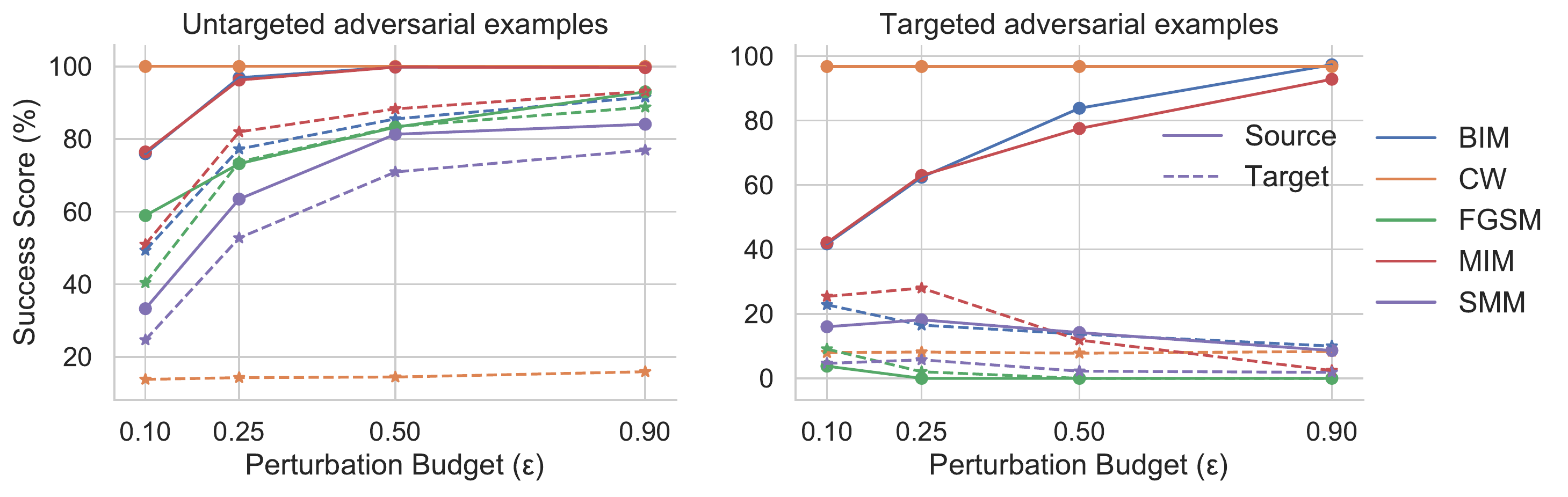}
    \caption{Success score of untargeted and targeted adversarial examples computed using the source system on source and target system for the MHEALTH dataset. The adversary has white-box access to the source system and black-box access to the target system. The adversary computes adversarial examples at different adversarial perturbation budgets using multiple evasion attack methods and the ``Walking'' activity is set as the target class for targeted attacks.}
    \label{fig:mh_trans_subjects}
\end{figure}

\begin{figure}[tbh!]
    \centering
    \includegraphics[width=\linewidth]{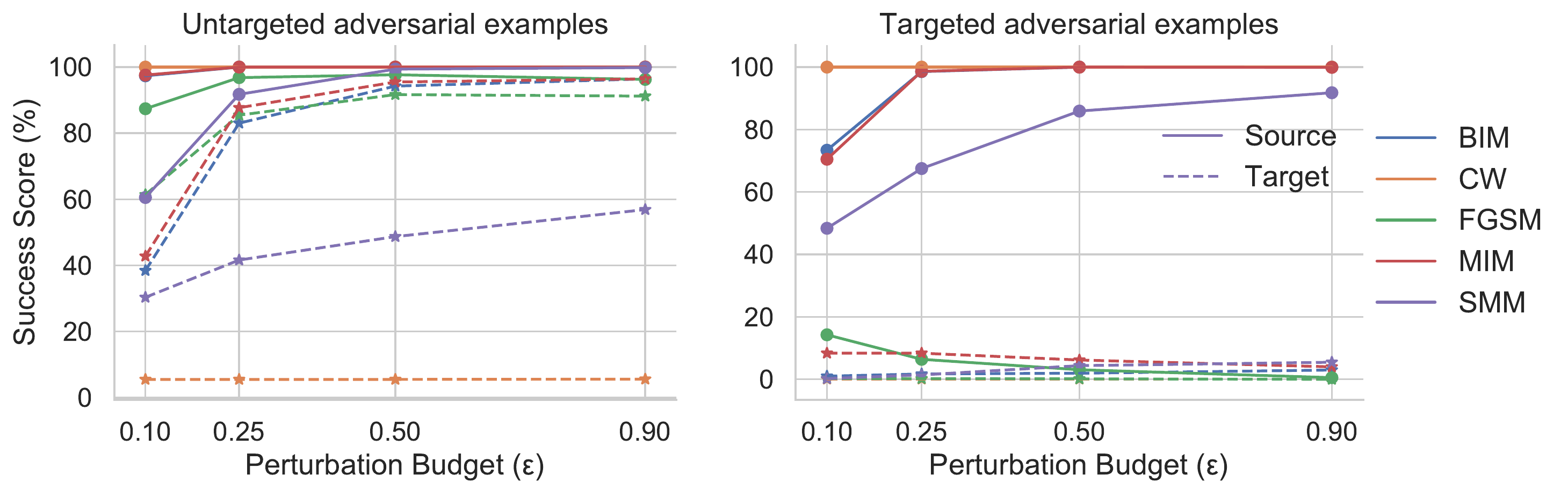}
    \caption{Success score of untargeted and targeted adversarial examples computed using the source system on source and target system for the UCI dataset. The adversary has white-box access to the source system and black-box access to the target system. The adversary computes adversarial examples at different adversarial perturbation budgets using multiple evasion attack methods and the ``Sitting'' activity is set as the target class for targeted attacks.}
    \label{fig:uci_trans_subjects}
\end{figure}

\begin{figure}[tbh!]
    \centering
    \includegraphics[width=\linewidth]{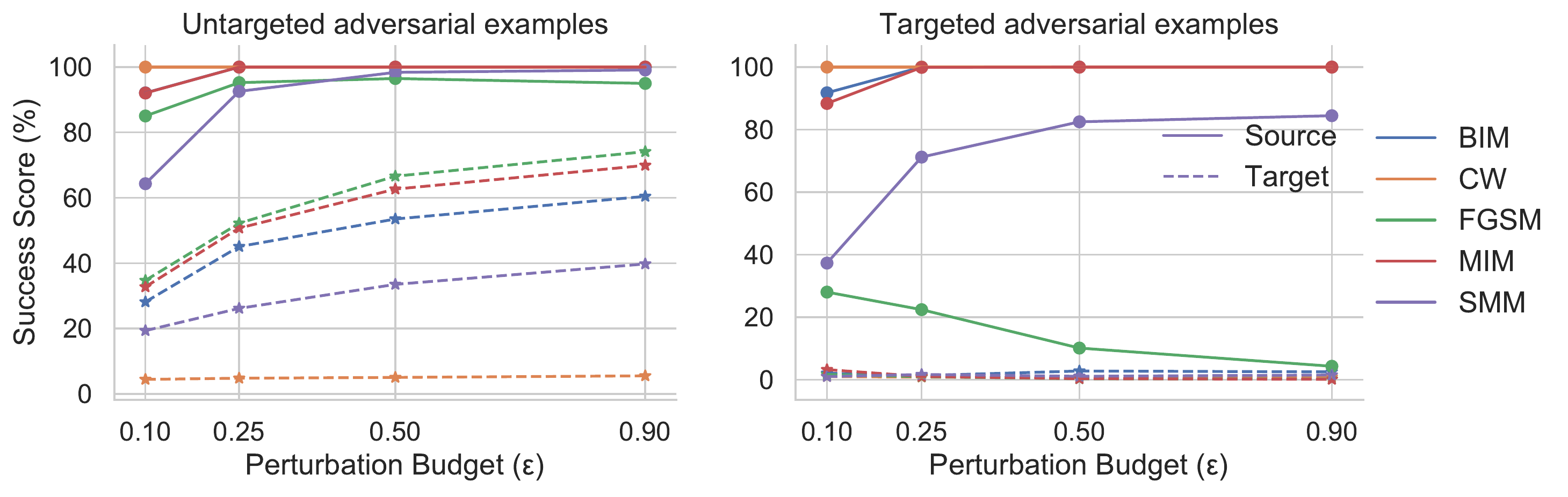}
    \caption{Success score of untargeted and targeted adversarial examples computed using the source system on source and target system for the DL dataset. The adversary has white-box access to the source system and black-box access to the target system. The adversary computes adversarial examples at different adversarial perturbation budgets using multiple evasion attack methods and the ``Sitting'' activity is set as the target class for targeted attacks.}
    \label{fig:dl_trans_subjects}
\end{figure}

Figures \ref{fig:mh_trans_subjects}, \ref{fig:uci_trans_subjects}, and \ref{fig:dl_trans_subjects} shows the success score of untargeted and targeted adversarial examples computed by the adversary using the source system on source and target systems for MHEALTH, UCI, and DL datasets. For targeted attacks, the activity class ``\textit{Walking}'' is used as the target class for the MHEALTH dataset and the activity class ``\textit{Sitting}'' is used as the target class for UCI and DL datasets. As expected, untargeted adversarial examples are highly transferable to the target system for all three datasets. But success score for the DL dataset is lower compared to the UCI and MHEALTH dataset, indicating low generalization we observed between the source and target system for the DL dataset. Targeted adversarial examples were unsuccessful for all three datasets, confirming that the noise and artifacts are specific to each user can greatly affect the adversarial transferability. 

\subsection{Adversarial Transferability Across Sensor Body Locations}
In wearable systems, sensors can be placed at different body positions to measure the same physiological change and biomarkers to detect physical activities. For example, to detect human activity, a person can use wearable devices that can be placed at different body positions like a wrist, ankle, waist, and chest. The placement of the sensor at different body locations introduces variations in sensor readings due to orientation and physical force experienced by sensor units. These variations result in the corresponding machine learning algorithms trained on the data from sensor placed at different body locations learn unique mappings between inputs and outputs and consequently transferability of adversarial examples will differ greatly from the traditional transferability of adversarial between machine learning models. In this section, we evaluate adversarial transferability across different sensor body locations for targeted and untargeted evasion attacks. The MHEALTH dataset has readings from accelerometers placed at three different body positions. The first sensor is wrapped around the subject chest, the second is worn by the subject on the right wrist, and the last one is worn on the left ankle. All the sensors have the same physical and electrical characteristics, and the subjects perform physical activities in the natural setting without any intervention and control measures. To evaluate adversarial transferability across sensor body locations, we perform experiments with different choice of the source and target system. In the first case, the data from the chest sensor is used to train the source system and the data from the ankle sensor is used to train the target system. Furthermore, in the second case, the data from the wrist sensor is used to train the source system and the data from the chest sensor is used in the target system. Finally, we have data from ankle sensor used in the source system and data from wrist sensor is used to train the target system. Just as we have done for the earlier cases of adversarial transferability, the adversary has complete access to the source system and no knowledge of the target system. Also, we use the same architecture of Convolutional Neural Networks (CNN), described for transferability across subjects, for the source and target models architecture.

After training the source and target model on their respective datasets obtained from sensors placed at different body locations, we evaluate the trained source and target models on both datasets. Table \ref{tab:source_target_clfs_acc_trans_body_locs} shows the classification accuracy of these models on both datasets for the three cases of source and target sensor body position. For all cases of adversarial transferability across sensor body locations the classification accuracy of the source model on the target dataset and the target model on the source dataset is very low, indicating low generalization between the source and target systems. 

\begin{table}[!htb]
    \caption{The classification accuracy of source and target systems on the source and target datasets for different cases of adversarial transferability across sensor body locations. For example, the table for \textbf{Chest - Ankle} shows the classification accuracy of the source system (trained on the data from Chest sensor) and target system (trained on the data from the left Ankle sensor) on both source and target datasets.}
    \label{tab:source_target_clfs_acc_trans_body_locs}
    \begin{tabular}{c}
        \toprule
        \multicolumn{1}{c}{Machine Learning} \\
        {System}\\
        \midrule
        Source \\
        Target \\
        \bottomrule
    \end{tabular}
    \quad
    \begin{tabular}{cc}
        \toprule
        \multicolumn{2}{c}{Chest Vs. Left-Ankle} \\
        \cline{1-2} 
        {Source Set} & {Target Set}\\
        \midrule
        98.81\% & 12.20\% \\
        18.75\% & 96.75\% \\
        \bottomrule
    \end{tabular}
    \quad
    \begin{tabular}{cc}
        \toprule
        \multicolumn{2}{c}{Right-Wrist Vs. Chest} \\
        \cline{1-2} 
        {Source Set} & {Target Set}\\
        \midrule
        99.67\% & 18.69\% \\
        23.58\% & 99.17\% \\
        \bottomrule
    \end{tabular}
    \quad
    \begin{tabular}{cc}
        \toprule
        \multicolumn{2}{c}{Left-Ankle Vs. Wrist} \\
        \cline{1-2} 
        {Source Set} & {Target Set}\\
        \midrule
        98.55\% & 22.06\% \\
        19.89\% & 99.43\% \\
        \bottomrule
    \end{tabular}
\end{table}

\subsubsection{Chest Vs. Left-Ankle}
\begin{figure}[!htb]
    \centering
    \includegraphics[width=\linewidth]{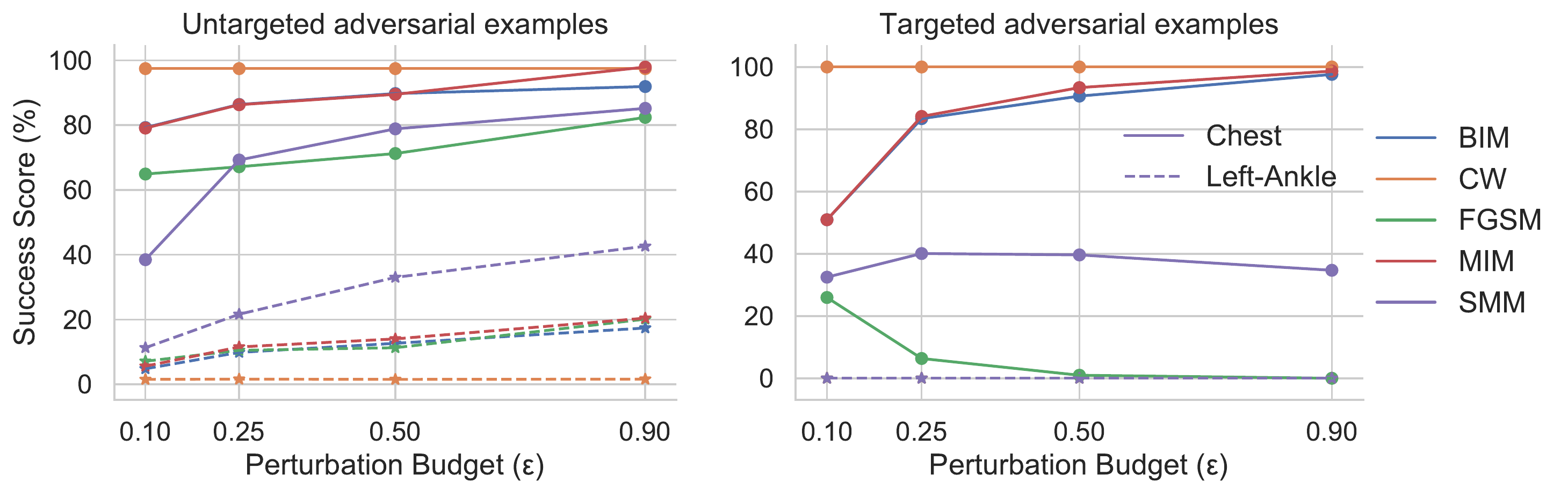}
    \caption{Success score of untargeted and targeted adversarial examples computed using the source system (\textbf{Chest}) on source and target (\textbf{Left-Ankle}) systems. The adversary has white-box access to the source system trained using the Chest dataset and black-box access to the target system trained on the Ankle dataset. The adversary computes adversarial examples at different adversarial perturbation budgets using multiple evasion attack methods and the ``Sitting'' activity is set as the target class for targeted attacks.}
    \label{fig:mh_trans_chest_ankle}
\end{figure}

Figure \ref{fig:mh_trans_chest_ankle} shows the success score of untargeted and targeted adversarial examples computed using the chest (source) system on the chest and left-ankle (target) system. We found good transferability for untargeted attacks and very low transferability for targeted attacks for target activity class of ``Sitting''. Untargeted adversarial examples with success score of $100\%$ on the source (chest) model performed poorly, success score in the range $0\% - 40\%$, on the target (left-ankle) model. The adversarial transferability further decrease for targeted attacks with adversarial success score of almost $0\%$ on the target system while the success score was in the range of $20\% - 100\%$ on the source system.

\subsubsection{Right-Wrist Vs. Chest}
\begin{figure}[!htb]
    \centering
    \includegraphics[width=\linewidth]{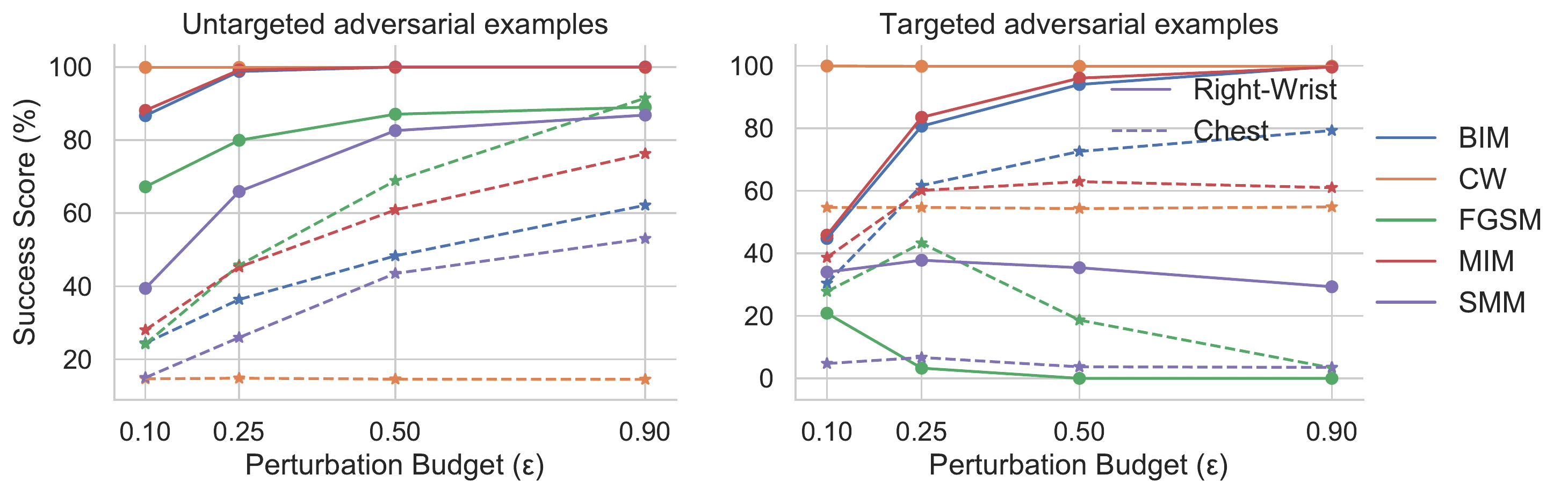}
    \caption{Success score of untargeted and targeted adversarial examples computed using the source system (\textbf{Right-Wrist}) on source and target (\textbf{Chest}) systems. The adversary has white-box access to the source system trained using the Wrist dataset and black-box access to the target system trained on the Chest dataset. The adversary computes adversarial examples at different adversarial perturbation budgets using multiple evasion attack methods and the ``Sitting'' activity is set as the target class for targeted attacks.}
    \label{fig:mh_trans_wrist_chest}
\end{figure}

Figure \ref{fig:mh_trans_wrist_chest} shows the success score of untargeted and targeted adversarial examples computed using the right-wrist (source) system on the right-wrist and chest (target) system. Surprisingly, we found high adversarial transferability for both untargeted and targeted attacks, with untargeted success score upto $> 90\%$ and targeted success score upto $80\%$ for the target system.

\subsubsection{Left-Ankle Vs. Right-Wrist}
\begin{figure}[!htb]
    \centering
    \includegraphics[width=\linewidth]{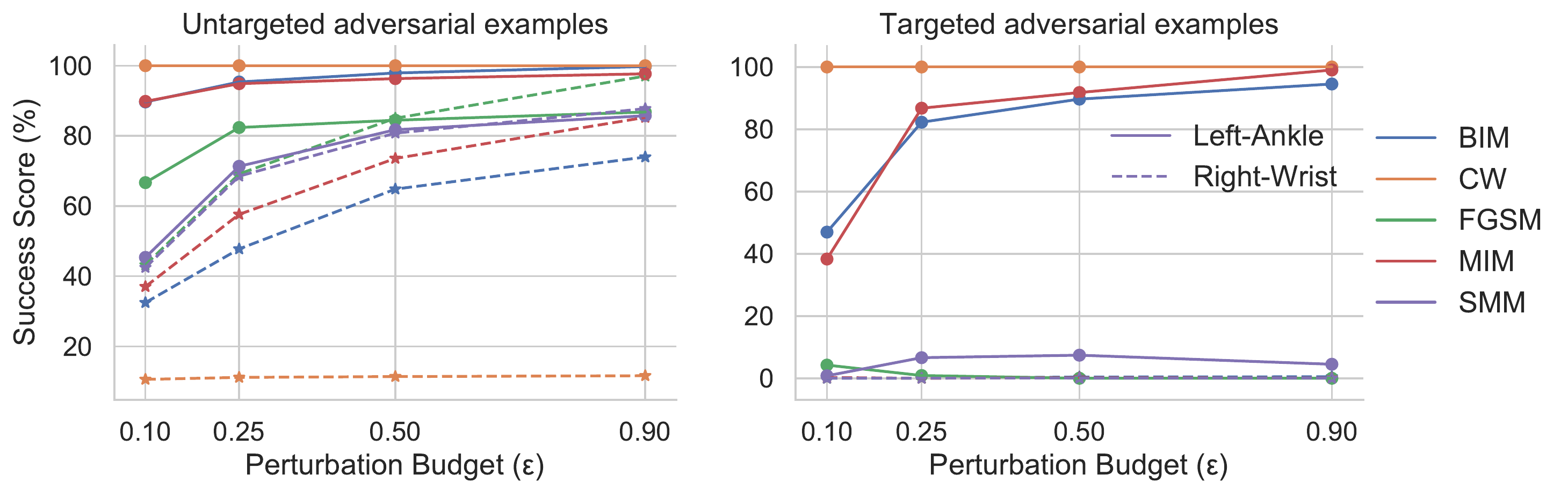}
    \caption{Success score of untargeted and targeted adversarial examples computed using the source system (\textbf{Left-Ankle}) on source and target (\textbf{Right-Wrist}) systems. The adversary has white-box access to the source system trained using the Ankle dataset and black-box access to the target system trained on the Wrist dataset. The adversary computes adversarial examples at different adversarial perturbation budgets using multiple evasion attack methods and the ``Sitting'' activity is set as the target class for targeted attacks.}
    \label{fig:mh_trans_ankle_wrist}
\end{figure}

Figure \ref{fig:mh_trans_ankle_wrist} shows the success score of untargeted and targeted adversarial examples computed using the left-ankle (source) system on the left-ankle and right-wrist (target) systems. Similar to Chest Vs. Left-Ankle case, we found high untargeted transferability, success score upto $98\%$, and very low $(0\%)$ targeted transferability.

The above results shows that adversarial transferability differe greatly with the sensor body location for the source and target systems. If the sensors for the source and target system are located near to each other, for example in the case of \textit{Right-Wrist Vs. Chest}, adversarial examples generated for the source system was able to fool the target system fairly for targeted and untargeted attacks. But for source and target sensor sensors that are placed far-apart on the body, in the case of \textit{Left-Ankle Vs. Chest} and \textit{Left-Ankle Vs. Right-Wrist}, the transferability of adversarial examples was low. Specifically, the success score of targeted adversarial examples was almost $0\%$ for all attack methods at all values of adversarial perturbation budget as shown in figures \ref{fig:mh_trans_ankle_wrist} and \ref{fig:mh_trans_chest_ankle}.

\subsection{Adversarial Transferability Between Datasets}
The final frontier of adversarial transferability in the context of wearable sensor systems is transferability across different datasets. Transferability between datasets includes all other types of transferability we have discussed so far and augments that with new variables such as sensor types, electrical properties of the sensor, data processing steps, and many more. To evaluate adversarial transferability between datasets, we train two CNN models, source and target, of same architecture and hyperparameters on different datasets. Since, we have three different datasets to evaluate adversarial transferability we have three different combinations for evaluation. Each of the combination assigns different dataset to the target and source systems. Once again the adversary has complete access to the source system and black-box access to the target system. In the first experiment, we assigned the UCI dataset to the source system and the MHEALTH dataset to the target system. Tables \ref{tab:uci_vs_mhealth_untar} and \ref{tab:uci_vs_mhealth_tar} shows the success score of untargeted and targeted adversarial examples for this case. In the second case, the DL dataset was assigned to the source system and the target system was trained on the UCI dataset. Tables \ref{tab:uci_vs_mhealth_untar} and \ref{tab:uci_vs_mhealth_tar} shows the success score of untargeted and targeted adversarial examples for the second case. Finally, in the third experiment the source system was trained on the MHEALTH dataset and the target system was trained on the DL dataset. We have not presented the result for the third case because the results were very similar to results for the first case of UCI source dataset and MHEALTH target dataset. We found poor adversarial transferability for the first and second cases, with highest success score of untargeted adversarial examples of $40.23\%$ and $0\%$ success score for targeted adversarial examples. The second case with DL source dataset and UCI target dataset, we observed untargeted success score up to $82\%$ and highest targeted success score of $11.51\%$. One interesting thing to note here is that targeted adversarial examples were more transferable at lowest adversarial perturbation budget $(\epsilon = 0.1)$ and untargeted adversarial examples were most transferable at highest value of adversarial perturbation budget $(\epsilon = 0.9)$.

\begin{table}[!tbh]
    \caption{Success score of untargeted adversarial examples computed using the source (UCI) system on the source and target (MHEALTH) systems.}
    \label{tab:uci_vs_mhealth_untar}
    \begin{tabular}{c}
        \toprule
        \multicolumn{1}{c}{Evasion} \\
        {Attack} \\
        {Methods} \\
        \midrule
        FGSM \\
        BIM \\
        MIM \\
        SMM \\
        CW \\
        \bottomrule
    \end{tabular}
    \quad
    \begin{tabular}{cc|cc|cc|cc}
        \toprule
        \multicolumn{8}{c}{Untargeted Attack Perturbation Budget ($\epsilon$)} \\
        \cline{1-8}
        \multicolumn{2}{c}{0.1} & \multicolumn{2}{c}{0.25} & \multicolumn{2}{c}{0.5} & \multicolumn{2}{c}{0.9} \\
        \cline{1-2} \cline{3-4} \cline{5-6} \cline{7-8}
        {Source} & {Target} & {Source} & {Target} & {Source} & {Target} & {Source} & {Target}\\
        \midrule
         74.09 & 0.11 & 84.15 & 0.85 & 87.65 & 11.96 & 89.74 & 36.15 \\
         86.91 & 0.19 & 96.89 & 0.62 & 97.86 & 8.03 & 97.94 & 25.78 \\
         86.79 & 0.19 & 96.07 & 0.97 & 96.50 & 13.09 & 96.07 & 40.23 \\
         51.84 & 0.11 & 78.17 & 0.11 & 92.73 & 0.11 & 96.07 & 0.11 \\
         100.0 & 0.03 & 100.0 & 0.03 & 100.0 & 0.03 & 100.0 & 0.03 \\
        \bottomrule
    \end{tabular}
\end{table}

\begin{table}[!tbh]
    \caption{Success score of targeted adversarial examples computed using the source (UCI) system on the source and target (MHEALTH) systems.}
    \label{tab:uci_vs_mhealth_tar}
    \begin{tabular}{c}
        \toprule
        \multicolumn{1}{c}{Evasion} \\
        {Attack} \\
        {Methods} \\
        \midrule
        FGSM \\
        BIM \\
        MIM \\
        SMM \\
        CW \\
        \bottomrule
    \end{tabular}
    \quad
    \begin{tabular}{cc|cc|cc|cc}
        \toprule
        \multicolumn{8}{c}{Targeted Attack Perturbation Budget ($\epsilon$)} \\
        \cline{1-8}
        \multicolumn{2}{c}{0.1} & \multicolumn{2}{c}{0.25} & \multicolumn{2}{c}{0.5} & \multicolumn{2}{c}{0.9} \\
        \cline{1-2} \cline{3-4} \cline{5-6} \cline{7-8}
        {Source} & {Target} & {Source} & {Target} & {Source} & {Target} & {Source} & {Target}\\
        \midrule
         8.62 & 0.0 & 3.82 & 0.0 & 1.02 & 0.0 & 0.55 & 0.0 \\
         64.83 & 0.0 & 92.39 & 0.0 & 99.02 & 0.0 & 99.95 & 0.0 \\
         58.58 & 0.0 & 91.18 & 0.0 & 99.95 & 0.0 & 100.0 & 0.0 \\
         45.10 & 0.0 & 49.02 & 0.0 & 52.0 & 0.0 & 35.68 & 0.0 \\
         99.95 & 0.0 & 99.95 & 0.0 & 99.95 & 0.0 & 99.95 & 0.0 \\
        \bottomrule
    \end{tabular}
\end{table}

\begin{table}[!tbh]
    \caption{Success score of untargeted adversarial examples computed using the source (DL) system on the source and target (UCI) systems.}
    \label{tab:dl_vs_uci_untar}
    \begin{tabular}{c}
        \toprule
        \multicolumn{1}{c}{Evasion} \\
        {Attack} \\
        {Methods} \\
        \midrule
        FGSM \\
        BIM \\
        MIM \\
        SMM \\
        CW \\
        \bottomrule
    \end{tabular}
    \quad
    \begin{tabular}{cc|cc|cc|cc}
        \toprule
        \multicolumn{8}{c}{Untargeted Attack Perturbation Budget ($\epsilon$)} \\
        \cline{1-8}
        \multicolumn{2}{c}{0.1} & \multicolumn{2}{c}{0.25} & \multicolumn{2}{c}{0.5} & \multicolumn{2}{c}{0.9} \\
        \cline{1-2} \cline{3-4} \cline{5-6} \cline{7-8}
        {Source} & {Target} & {Source} & {Target} & {Source} & {Target} & {Source} & {Target}\\
        \midrule
         69.53 & 11.80 & 91.92 & 27.88 & 94.42 & 51.20 & 90.84 & 82.52 \\
         91.92 & 7.08 & 99.48 & 13.75 & 99.48 & 33.43 & 99.48 & 55.09 \\
         93.01 & 8.51 & 99.48 & 20.32 & 99.48 & 41.32 & 99.48 & 74.73 \\
         51.47 & 25.28 & 83.96 & 38.69 & 93.59 & 41.98 & 95.27 & 48.50 \\
         100.0 & 5.47 & 100.0 & 5.59 & 100.0 & 5.84 & 100.0 & 6.01 \\
        \bottomrule
    \end{tabular}
\end{table}

\begin{table}[!tbh]
    \caption{Success score of targeted adversarial examples computed using the source (DL) system on the source and target (UCI) systems.}
    \label{tab:dl_vs_uci_tar}
    \begin{tabular}{c}
        \toprule
        \multicolumn{1}{c}{Evasion} \\
        {Attack} \\
        {Methods} \\
        \midrule
        FGSM \\
        BIM \\
        MIM \\
        SMM \\
        CW \\
        \bottomrule
    \end{tabular}
    \quad
    \begin{tabular}{cc|cc|cc|cc}
        \toprule
        \multicolumn{8}{c}{Targeted Attack Perturbation Budget ($\epsilon$)} \\
        \cline{1-8}
        \multicolumn{2}{c}{0.1} & \multicolumn{2}{c}{0.25} & \multicolumn{2}{c}{0.5} & \multicolumn{2}{c}{0.9} \\
        \cline{1-2} \cline{3-4} \cline{5-6} \cline{7-8}
        {Source} & {Target} & {Source} & {Target} & {Source} & {Target} & {Source} & {Target}\\
        \midrule
         0.43 & 11.51 & 0.17 & 4.03 & 0.0 & 4.34 & 0.0 & 2.04 \\
         52.24 & 11.70 & 67.64 & 9.32 & 90.62 & 5.81 & 97.02 & 4.17 \\
         56.22 & 7.41 & 71.52 & 3.90 & 98.30 & 5.09 & 99.97 & 5.25 \\
         4.05 & 2.26 & 3.73 & 1.07 & 1.92 & 0.11 & 0.67 & 0.05 \\
         99.85 & 1.43 & 99.85 & 1.46 & 99.82 & 1.41 & 99.82 & 1.46 \\
        \bottomrule
    \end{tabular}
\end{table}

\section{Discussion}
In this section, we discuss our results in detail and provide theoretical and graphical explanations. We also generate adversarial examples using non-parametric machine learning algorithms such as Decision Tree Classifier (DTC) and K-Nearest Neighbor classifier (KNN) and measure adversarial transferability on different learning algorithms. We also discuss adversarial transferability through the lens of manifold learning and non-robust features to provide intuition for our results and establish ideas for future research. 

\subsection{Adversarial Attacks with Decision Trees and K-Nearest Neighbors}
In our analysis, on adversarial transferability between machine learning models, we found KNN and DTC classifiers to be robust against targeted adversarial examples computed using a Deep Neural Network (DNN) compared to othe learning algorithms such as Support Vector Classifier (SVC) and Linear Regression Classifier (LRC). To further evaluate the robustness of non-parametric learning algorithms such as KNN and DTC, we computed targeted and untargeted adversarial examples using the KNN and DTC at the adversarial perturbation budget of $\epsilon = 0.5$. We used the Region Based Attack (RBA) \cite{yang2020robustness} and heuristic Decision Tree attack (Papernot) \cite{papernot2016practical} to compute adversarial examples using the Decision Tree Classifier. Region based attack find the closet polyhedron to an input where the classifier predicts a label other than the actual label and outputs the closest point in this region as an adversarial example. Region based attack is optimal and can find highly successful adversarial examples but suffer from high computational load. Heuristic Decision Tree attack is computationally very fast and simply searches for leaves in the decision tree with different class in the neighborhood of the leaf corresponding to the decision tree's original prediction for an input. The path from the original leaf to the adversarial leaf is used to modify the input sample to create an adversarial example. Kernel Substitution Attack \cite{papernot2016practical}, which uses the Fast Gradient Sign Method (FGSM) to craft adversarial examples misclassified by nearest neighbors, is used to compute untargeted adversarial examples for the KNN model. 

\begin{figure}[tbh!]
    \centering
    \includegraphics[width=\linewidth]{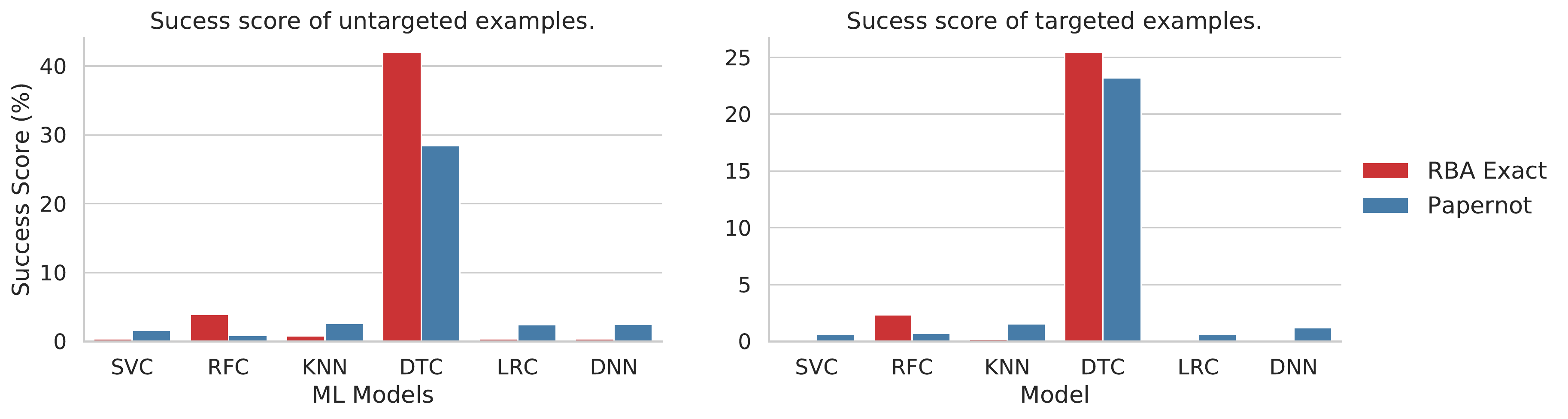}
    \caption{Success score of untargeted and targeted adversarial examples computed using the Decision Tree Classifier (DTC) on different machine learning models. Both untargeted and targeted adversarial examples were computed at the perturbation budget $\epsilon = 0.5$ and the target class was set to ``Sitting'' for targeted examples. We used the RBA and Papernot attacks to compute adversarial examples using the DTC.}
    \label{fig:uci_trans_model_dtc}
\end{figure}

\begin{figure}[tbh!]
    \centering
    \includegraphics[width=\linewidth]{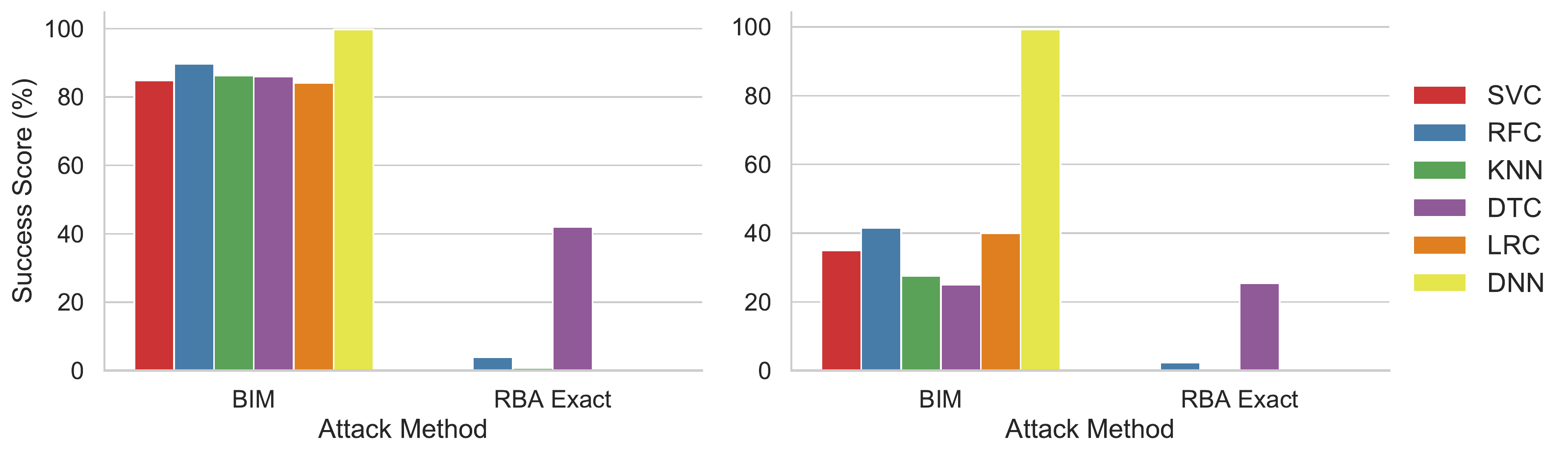}
    \caption{Success score of untargeted and targeted adversarial examples computed using the Decision Tree Classifier (DTC) and Deep Neural Network (DNN) on different machine learning models. Both untargeted and targeted adversarial examples were computed at the perturbation budget $\epsilon = 0.5$ and the target class was set to ``Sitting'' for targeted examples. We used the RBA method to generate adversarial examples using the DTC model and Basic Iterative Method (BIM) to generate adversarial examples using the DNN model.}
    \label{fig:uci_trans_model_dtc_dnn}
\end{figure}

Similar to the case of adversarial transferability between machine learning models, we trained $6$ different machine learning algorithms on the feature data of the UCI dataset, and computed adversarial examples using the test set with DTC and KNN models. Figure \ref{fig:uci_trans_model_dtc}, shows the success score of untargeted and targeted adversarial examples computed using the DTC model for all $6$ machine learning models. Adversarial examples were able to fool the DTC model with good success score $(40\%)$, but performed poorly on other models, indicating poor adversarial transferability in both targeted and untargeted cases. We also want to highlight the difference in the success score of adversarial examples computed using the DNN model, in section \ref{section:adv_tr_bet_models} and adversarial examples computed using the DTC model here. Figure \ref{fig:uci_trans_model_dtc_dnn} shows the success score of adversarial examples computed using the DTC with RBA method and DNN model using Basic Iterative Method (BIM) on all $6$ machine learning models. As we can see, adversarial examples computed using the DNN model are more successful on the DTC, compared to adversarial examples computed using the DTC for the same adversarial perturbation budget. Furthermore, adversarial examples computed using the DNN model are highly/more transferable than adversarial examples computed using the DTC, indicating the importance of adversarial transferability.

\begin{table}[tbh]
  \caption{Success score of untargeted adversarial examples computed using the k-Neareat Neighbor (KNN) model on different machine learning models for the adversarial perturbation budget of $\epsilon = 0.5$. The Kernel Substitution attack method was used to compute adversarial examples using the KNN model.}
  \label{tab:uci_trans_model_knn}
  \begin{tabular}{ccccccc}
     & \multicolumn{6}{c}{Machine Learning Models} \\
    \cline{2-7}
    & {SVC} & {RFC} & {KNN} & {DTC} & {LRC} & {DNN} \\
    \midrule
    Success Score (\%) & 85.74 & 89.82 & 86.01 & 85.32 & 86.40 & 77.35 \\
  \bottomrule
\end{tabular}
\end{table}

Table \ref{tab:uci_trans_model_knn} shows the success score of untargeted adversarial examples computed using the KNN model for all $6$ machine learning models. The adversarial examples are highly transferable and the success score are similar to that obtained with adversarial examples generated using the DNN as shown in section \ref{section:adv_tr_bet_models}. Therefore, KNN model is more vulnerable compared to the DTC model at the same level of adversarial perturbation budget. But adversarial attack methods that works with non-parametric learning algorithms such as DTC and KNN are much more computational intensive compared to gradient based adversarial attack method and adversarial examples computed with non-parametric attack methods are also less successful and transferable. Hence, with computation and resource limitation that are inherent to sensor systems, a direct attack on non-parametric learning algorithms might not be feasible and it makes better sensor for an attacker to use gradient based attack methods to compute adversarial examples using the source systems and attack the target system by exploiting the transferability of adversarial examples.

\label{feature_overlap}
\subsection{Feature Overlap}
Authors in \cite{NIPS2019_8307}, have argued that neural networks trained on independent samples from a distribution tend to learn similar ``\textit{non-robust}'' or brittle features making adversarial transferability possible. The central thesis to consider is here is that data samples used to train the machine learning model and used by an adversary belong to the same distribution. Therefore, in theory, models trained on similar data distributions have strong transferability between them, and models trained on distinct data distributions have weak transferability. This is because similar data distributions facilitate the learning of similar non-robust features, and different data distribution has minimal overlap between the corresponding non-robust features. For different adversarial transferability cases we have analyzed in this work, the data distribution of the source and the target models has a varying degree of similarity. In particular, for transferability between models, all models are trained on the same dataset. In this case, different models are trained on the same dataset and learn similar non-robust features, demonstrating excellent adversarial transferability. On the other hand, for transferability between datasets, the source and target model are trained on datasets from different distributions, making it very hard for the models to learn similar non-robust features, and resulting in poor transferability in this case. To verify this, we evaluated the target model on the test set of the source model. The performance of the target model on the source model's test set in theory is correlated with learned features shared between them. Higher classification accuracy of the target model on the source model test set means learning of similar features, and lower classification accuracy demonstrates learning of different features between the source and the target model. And the degree to which the target and source model share learned features is proportional to the performance of the target model on the adversarial examples computed using the source model. 

\begin{figure}[!tbh]
    \centering
    \includegraphics[width=0.8\textwidth]{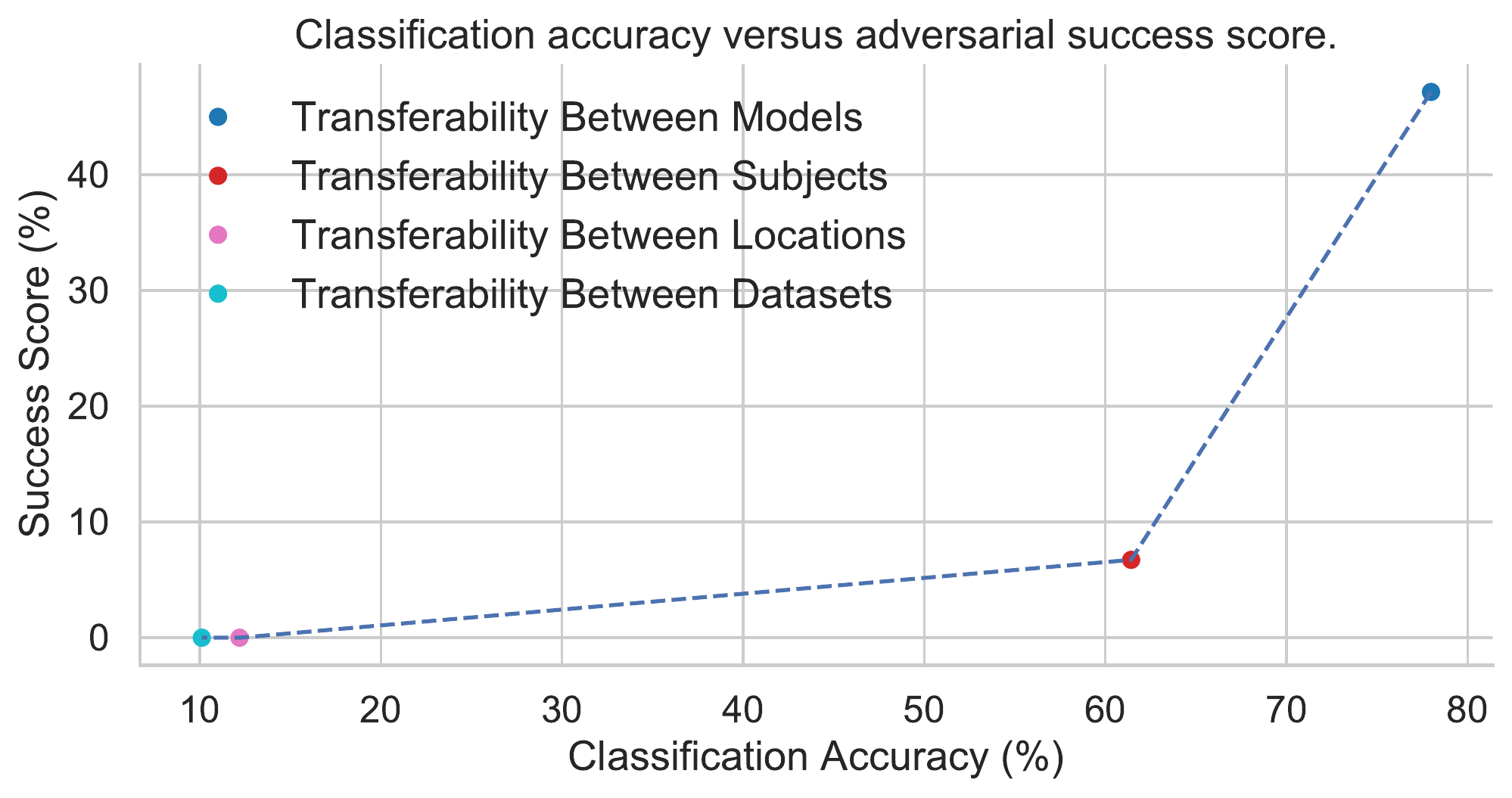}
    \caption{The classification accuracy of the target model for different cases of transferability on the test set of the source model vs. the success score of targeted adversarial examples computed using the Basic Iterative Method (BIM) on the target model.}
    \label{fig:non_robust_features}
\end{figure}

Fig \ref{fig:non_robust_features} shows the classification accuracy and success score of the target model on the source model test set and targeted adversarial examples computed using the source model with the Basic Iterative Attack (BIM). As we can see, the performance of the target model on the source model test set is directly proportional to the target model's performance on adversarial examples. Higher classification accuracy on the test set corresponds to a higher success score on adversarial examples and vice-versa. This demonstrates that the degree to which features are shared between the target and source model is directly related to the effectiveness of adversarial examples on the target model. Hence, learning of similar features by target and source models facilitates better adversarial transferability, as demonstrated in transferability between models and transferability between subjects. On the other hand, when the target and source model have less overlap between learned features, the adversarial transferability is also poor.

\subsection{Manifold Learning}
Manifold learning methods seek to describe high-dimensional data in low dimensional space. ``\textit{The central idea underlying these methods is that although natural data is typically represented in very high-dimensional spaces, the process generating the data is often thought to have relatively few degrees of freedom}'' \cite{Zheng2009}. Manifold learning allows us to generate low-dimensional embedding of the data, which can reveal interesting patterns not possible to observe and represent in its high dimensional form.  Manifold learning relies on the distance between each point and other points in the dataset to get the low-dimensional representation. We will use \textit{Multidimensional Scaling} (MDS) \cite{kruskal_multidimensional_2020} to generate the low-dimensional embedding of adversarial examples and clean samples. Multidimensional scaling uses a pair-wise distance matrix as inputs and aims at placing each data point in an n-dimensional space such that the distance between the points is preserved as well as possible. Albeit, the Euclidean distance suffers from the curse of dimensionality when used to compute the distance between objects in high-dimensional space, it can still be used to compute the similarity matrix between adversarial and clean samples. This similarity matrix is used by multidimensional scaling to get the low-dimensional representation of the adversarial and clean samples. 

\begin{figure}[!tbh]
    \centering
    \includegraphics[width=\textwidth]{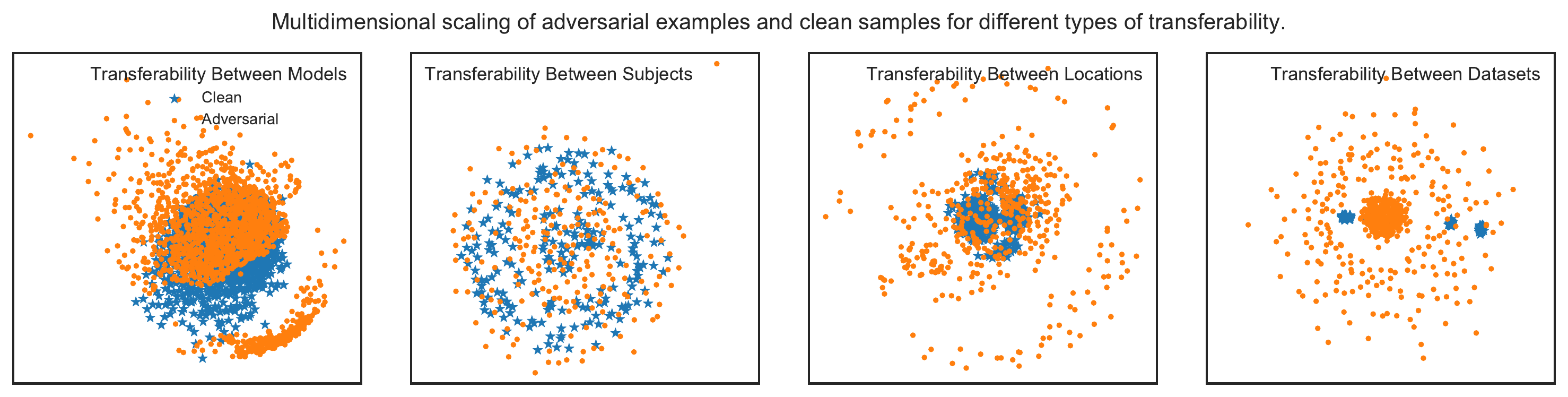}
    \includegraphics[width=\textwidth]{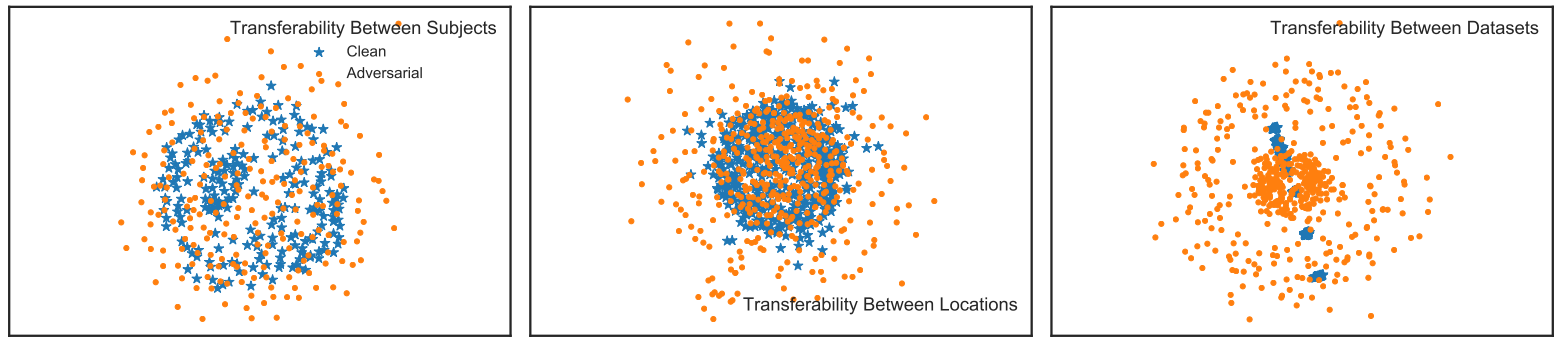}
    \includegraphics[width=\textwidth]{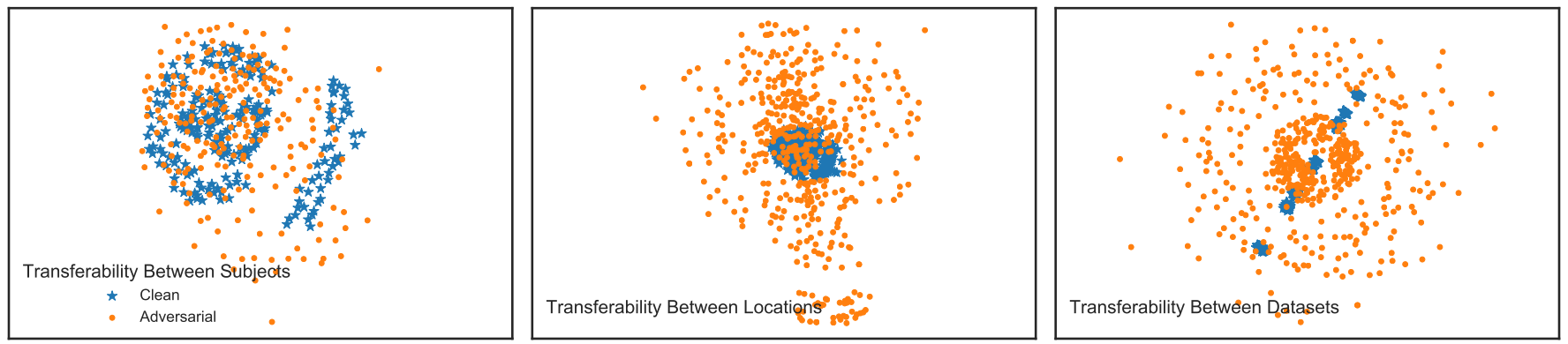}
    \caption{Multidimensional Scaling of clean and adversarial samples for different cases of targeted adversarial transferability. Except for transferability between models that work with $1$D feature data, other types of adversarial transferability uses $3-$axial accelerometer data and hence we have plots of $Y$ axis and $Z$ axis for these modes of adversarial transferability.}
    \label{fig:mds}
\end{figure}

For this experiment, we used targeted adversarial examples computed using the source model with the basic iterative attack and clean samples for the target class from the target model's training set. Fig \ref{fig:mds} shows the scatter plot of the $2-$dimensional embedding for different types of adversarial transferability we have discussed in this work. To compute the low dimensional embedding, we used all samples from the target model's training set for the target class and the top-$k$ samples from the adversarial set that was classified into the target class by the target model. Here, $k$ is the number of samples selected from the target model training set, and we sort the prediction confidence of the adversarial examples for the target class to determine the top-$k$ examples. In cases where adversarial examples fail to fool the target model, we take $k$ random samples from the adversarial set.

Fig \ref{fig:mds} shows the multidimensional scaling of adversarial examples and benign samples from the target model's training set for different cases of adversarial transferability. For transferability between models, the 2-dimensional representation of clean and adversarial samples share a significant overlap region, which corresponds to the high targeted transferability we observed in this case. The region of overlap for transferability across subjects is not significant, but the spatial distribution of adversarial and benign samples share shape and organization, which demonstrate the fair transferability in this case. For transferability across sensor body locations and datasets, the representation of benign and adversarial samples shares neither region nor organization and consequently we observed poor targeted adversarial transferability for these cases in our results. The representation of clean and adversarial samples in $2-$dimensional space gives us insights about the transferability results we obtained in our experiments, but we need to be careful when we represent high-dimensional objects in low-dimensional space. Our aim here was to demonstrate how the spatial distribution of adversarial and benign samples looks like for different cases of adversarial transferability and explain the results we obtained from our experiments. The degree to which adversarial samples can conform to benign samples from the target model's training set is directly proportional to adversarial examples success score on the target system.

\section{Conclusions}
Adversarial examples are shown to be transferable across machine learning models trained on the whole or subset of the same dataset. However, the problem of adversarial transferability does not end there. For the first time in literature, we have investigated novel types of adversarial transferability in the context of wearable sensor systems with an extensive set of experiments. These new aspects of adversarial transferability show how an adversary can exploit sensor systems properties to craft adversarial examples in ways not discussed before. Our results not only demonstrate that there exist many new types of adversarial transferability but also show where these techniques excel and fail. 

In our experiments, we have first discussed the transferability across machine learning models. Using the feature data from three real-world datasets, we found very strong untargeted transferability between different types of machine learning models with five attack methods. For targeted attacks, adversarial examples were less transferable for all three datasets and we found Decision Tree Classifier (DTC) and k-Nearest Neighbors (KNN) to be more robust compared to other types of learning algorithms. Also, the level of targeted transferability differed greatly across the three datasets. For Daily-Living (DL) dataset, both targeted and untargeted adversarial examples were much less likely to be transferable compared to the UCI and MHEALTH datasets. The underlying reasons behind the low adversarial transferability for the DL datasets, remains unclear but we believe the nature of the DL datasets makes it harder for an adversary to compute transferable adversarial examples. 

For cross subject transferability, we randomly selected data from half the subjects to create the source dataset and the data from the remaining half subjects was used as the target dataset. We separated all three datasets in the above manner, and presented results for cross subject transferability for all datasets. We discovered that the level of generalization between the source and target models, greatly affected the transferability of adversarial examples. For UCI and MHEALTH dataset, we found high level of generalization as indicated by the classification accuracy of the source model on the target dataset and the target model on the source data and consequently untargeted adversarial examples were highly transferable. For the DL dataset, we found low level of generalization between the source and target system and as a result low level of untargeted adversarial transferability. Targeted transferability was low for all three datasets, but for DL dataset the success score of targeted adversarial examples was $0\%$ even at higher values of adversarial perturbation budget.

In the next experiment, we evaluated the transferability across sensor body positions using the data from sensors placed at chest, left-ankle, and right-wrist. The adversarial transferability differed greatly based on the source and target system sensor body locations. For source and target system sensors that were located near to each other, for examples right-wrist and chest, adversarial examples generated using the source system were highly transferable to the target system for both targeted and untargeted attacks. But for source and target systems sensor that were placed far-part, for example left-ankle and chest, the transferability of targeted adversarial examples was low. Finally, in the last experiment, we analyzed transferability across datasets. For this, we used all three datasets: UCI, DL, and MHEALTH. All datasets have some common activity classes between them and some unique activity classes. For different combinations of source and target datasets, we found very low untargeted and targeted adversarial transferability for the entire spectrum of analysis.


In this work, we have explored novel directions of adversarial transferability in the context of sensor systems and showed how an adversary's performance varies in different transferability settings. In particular, our findings can be summarized as follows.

\begin{enumerate}
    \item The traditional notion of transferability - transferability across machine learning models - showed excellent results which is consistent with the literature. But we also discovered that the properties of the underlying data distribution greatly affects the success score of adversarial transferability.
    
    \item Non-parametric learning algorithms such as decision tree and k-nearest neighbor were more robust to targeted and untargeted adversarial examples computed using both gradient and non-gradient based attack methods. 
    
    \item We found gradient-based attack methods to be more competent at finding adversarial perturbations and were able to fool the target system for both targeted and untargeted evasion attacks. Attacks involving complicated optimization with significantly longer run times performed poorly in terms of transferability but showed higher misclassification and success score on the target model, even at lower epsilon values.
    
    \item In all cases of transferability except transferability between datasets, untargeted attacks were more successful than targeted attacks. This is because untargeted attacks are considered successful as long as they can achieve random misclassification, which is much easier to achieve. The nature of the time-series input to sensor systems makes them more vulnerable to random misclassification because the data they operate on have properties that are easier to exploit. However, the complexity of adversarial attacks increases significantly in the targeted case. For the targeted attacks, the attack methods have to find adversarial perturbations that need to conform to the temporal and spatial properties present in the dataset for the chosen target class. Due to this, the targeted transferability was very poor in most cases. Hence, targeted attacks give a better measure of adversarial transferability in general for all the different cases considered in this work.
\end{enumerate}

\section{Recommendations}
In this section, we provide recommendations aimed at a system designer for designing a robust wearable sensor systems based on our results and findings. These recommendations can be considered as design choices that can affect the adversarial robustness of a sensor system.

\begin{enumerate}
    \item We found the fundamental reason for the robustness of target system against untargeted and targeted adversarial examples computed using the source system was the distance between the source and target data distributions. With increasing level of distance between the source and target data distribution as shown in section \ref{feature_overlap}, the adversarial transferability of adversarial examples decreased and reached success score of $0\%$ for adversarial transferability between datasets. Hence, when designing and developing wearable sensor system it makes sense to use proprietary or private datasets. Also, if it is not possible to use private dataset, data processing techniques such as Principal Component Analysis \cite{bhagoji2017enhancing}, removing non-robust samples from the dataset \cite{yang2020robustness} and noise addition should be used as a preprocessing steps on the dataset to learn a robust classifier.
    
    \item Non-parametric machine learning algorithms that are local classifiers whose output depends on training data close to the test instance such as K-Nearest Neighbors (KNN) and Decision Tree Classifier (DTC) were found to be more robust against adversarial transferability compared to other types of learning algorithms. We further explored the robustness of non-parametric methods by generating adversarial examples using the KNN and DTC models. For the same level of adversarial perturbation budget, the adversarial examples computed using the non-parametric algorithms were less successful on the source system and less transferable to the target system compared to adversarial examples generated using a Deep Neural Network. Also, computing adversarial examples using non-parametric learning algorithms is more computationally intensive and may not feasible for resource constraints system such as wearable sensor systems. Hence, if possible we recommend to adopt non-parametric learning algorithms for wearable sensor systems.
    
    \item Sensor system trained on a large real-world dataset was discovered to be more robust to adversarial transferability compared to system trained with smaller lab-setting datasets. In our analysis, target systems that used the Daily Living (DL) dataset (sample size $16434$) were more robust towards both untargeted and targeted adversarial examples than target system that used the UCI dataset (sample size $10299$) and the MHEALTH dataset (sample $5133$). Hence, in practice it is better to have a large dataset for a robust system from adversarial transferability point of view.
    
\end{enumerate}

Finally, we want to draw the reader's attention to the argument that machine learning systems can be protected by access control, and very few cases of adversarial attacks can happen in real-world wearable systems. However, at the same time, we also want to underscore the fact that by limiting our understanding of vulnerabilities that exists in learning algorithms and models used in sensor systems by operating on the default setting that adversarial attacks on sensor systems have a low chance of occurrence is not practical. The results presented in this paper draws its strength not only from its applicability in real-world systems but also in its ability to show the vulnerability of wearable sensor systems and the lack of robustness of learning algorithm that an adversary can exploit to attack these systems. If we ignore the discussion of adversarial attacks and transferability by operating on the default setting, we will be blind to the inherent shortcomings of our systems, which can be detrimental to the overall health of our systems. For example, consider a fall detection system employed to dispatch help when the system detects falls. If an adversary can influence any aspect of this system, then the effect can have life-altering consequences. Furthermore, recent works have shown that adversarial attacks are possible in real-world conditions, and the transferability of adversarial examples dramatically enhances the chances of success for an adversary \cite{10.1145/3376897.3377856, ilyas2018black, gong2019real, goodman2020transferability}. Furthermore, the decision-making model needs not be present locally on the wearable sensor system. The model can be on the cloud, and the sensor systems operate by querying the cloud model with sensor readings for classification \cite{cruz2018automated, 8719325}. This mode of operation is becoming more mainstream as it provides many benefits, such as life-long learning, active learning, and data analytics. Therefore, acknowledging and understanding the adversarial nature of machine learning algorithms used in wearable systems allow us to build measures and adapt the design process to thwart and limit the impact of adversarial attacks. This is precisely what we aimed to achieve in this work. By making the connection between adversarial transferability and different aspects of wearable sensor systems, we showed where the strengths and limitations of an adversary lie and how a system designer can use this information to design and implement robust learning algorithms.

\section{Limitations and Future Works} 
In this work, we have tried to cover the topic of adversarial transferability in wearable sensor systems in a broad manner. Nonetheless, our work does have some limitations, which we have highlighted below.

\begin{itemize}
    \item In our experiments, we have used five different adversarial attack methods to evaluate adversarial transferability in wearable sensor systems. However, there are many more attack methods in the literature that we have left out of our discussion. Unexplored attack methods with better optimization strategy may be able to find adversarial perturbation with better transferability properties and succeed where the discussed attack methods have failed.
    
    \item The discussion of adversarial machine learning is not complete without talking about defense against adversarial attacks. Attack and defense form the two faces of the adversarial machine learning coin, and hence should be given equal importance and attention in research. Our discussion in this work does not discuss defense mechanisms, and we aim to explore the effects of defense methods against adversarial transferability in our future works. 
\end{itemize}

Based on the limitations of our work and results from the experiments, some possible research work that can build on our findings are:
\begin{itemize}
    \item In this work, we have only discussed the level of performance of different attack methods in terms of transferability. One interesting question that we can ask based on our results is, ``What makes some attack methods to have higher or lower rates transferability than others?''. We think this is one of the fundamental questions that need to be answered to better understand the results we have showed in this work.
    
    \item One straightforward extension to our work can address the limitations we have discussed earlier. This will involve including more types of attack methods for analysis in the framework we have outlined and also introduce defense methods in the picture.
    
\end{itemize}



\bibliographystyle{ACM-Reference-Format}
\bibliography{ms}

\appendix

\end{document}